%% file: main.tex
\documentclass[10pt,twocolumn,letterpaper]{article}
\usepackage[pagenumbers]{cvpr} % To force page numbers, e.g. for an arXiv version


\definecolor{myblue}{rgb}{0.21,0.49,0.74}
\usepackage[pagebackref,breaklinks,colorlinks,allcolors=myblue]{hyperref}
\usepackage{listings}
\usepackage{algorithm}
\usepackage{algpseudocode}

\definecolor{codegreen}{rgb}{0,0.6,0}
\definecolor{codegray}{rgb}{0.5,0.5,0.5}
\definecolor{codepurple}{rgb}{0.58,0,0.82}
\definecolor{backcolour}{rgb}{0.95,0.95,0.92}

\title{CAPE: A CLIP-Aware Pointing Ensemble of Complementary Heatmap Cues for Embodied Reference Understanding}

\author{Fevziye Irem Eyiokur\textsuperscript{1,4} \qquad Dogucan Yaman\textsuperscript{1,4} \qquad Hazım Kemal Ekenel\textsuperscript{2} \qquad Alexander Waibel\textsuperscript{1,3,4} \\
\textsuperscript{1}Karlsruhe Institute of Technology, \textsuperscript{2}Istanbul Technical University, \textsuperscript{3}Carnegie Mellon University \\
\textsuperscript{4}KIT Campus Transfer GmbH (KCT) \\
{\tt\small irem.eyiokur@kit.edu}}

\begin{document}
\maketitle
\input{sec/0_abstract}    
\input{sec/1_intro}
\input{sec/2_related_work}

\input{sec/3_method}
\input{sec/4_experiments}
\input{sec/5_conclusion}
{
    \small
    \bibliographystyle{ieeenat_fullname}
    \bibliography{main}
}

\clearpage
\renewcommand{\thesection}{\Alph{section}}
\renewcommand{\thefigure}{\Alph{section}.\arabic{figure}}
\renewcommand{\thetable}{\Alph{section}.\arabic{table}}
\renewcommand{\thealgorithm}{\Alph{section}.\arabic{algorithm}}
\appendix
\input{sec/X_supp}

\end{document}

%% file: sec/0_abstract.tex
\begin{abstract}
We address Embodied Reference Understanding, the task of predicting the object a person in the scene refers to through pointing gesture and language. This requires multimodal reasoning over text, visual pointing cues, and scene context, yet existing methods often fail to fully exploit visual disambiguation signals. We also observe that while the referent often aligns with the head-to-fingertip direction, in many cases it aligns more closely with the wrist-to-fingertip direction, making a single-line assumption overly limiting. To address this, we propose a dual-model framework, where one model learns from the head-to-fingertip direction and the other from the wrist-to-fingertip direction. We introduce a Gaussian ray heatmap representation of these lines and use them as input to provide a strong supervisory signal that encourages the model to better attend to pointing cues. To fuse their complementary strengths, we present the CLIP-Aware Pointing Ensemble module, which performs a hybrid ensemble guided by CLIP features. We further incorporate an auxiliary object center prediction head to enhance referent localization. We validate our approach on YouRefIt, achieving 75.0 mAP at 0.25 IoU, alongside state-of-the-art CLIP and $C_D$ scores, and demonstrate its generality on unseen CAESAR and ISL Pointing, showing robust performance across benchmarks.
\end{abstract}

%% file: sec/1_intro.tex
\section{Introduction}
\label{sec:intro}

Embodied Reference Understanding (ERU)~\cite{chen2021yourefit} is the task of identifying a specific object in a visual scene based on language instructions and pointing cues in the image. 
It plays a key role in real-world applications like human-robot interaction, assistive robotics, and augmented reality, where systems must understand which object a person refers to.

\begin{figure}
    \centering
    \includegraphics[width=0.82\linewidth]{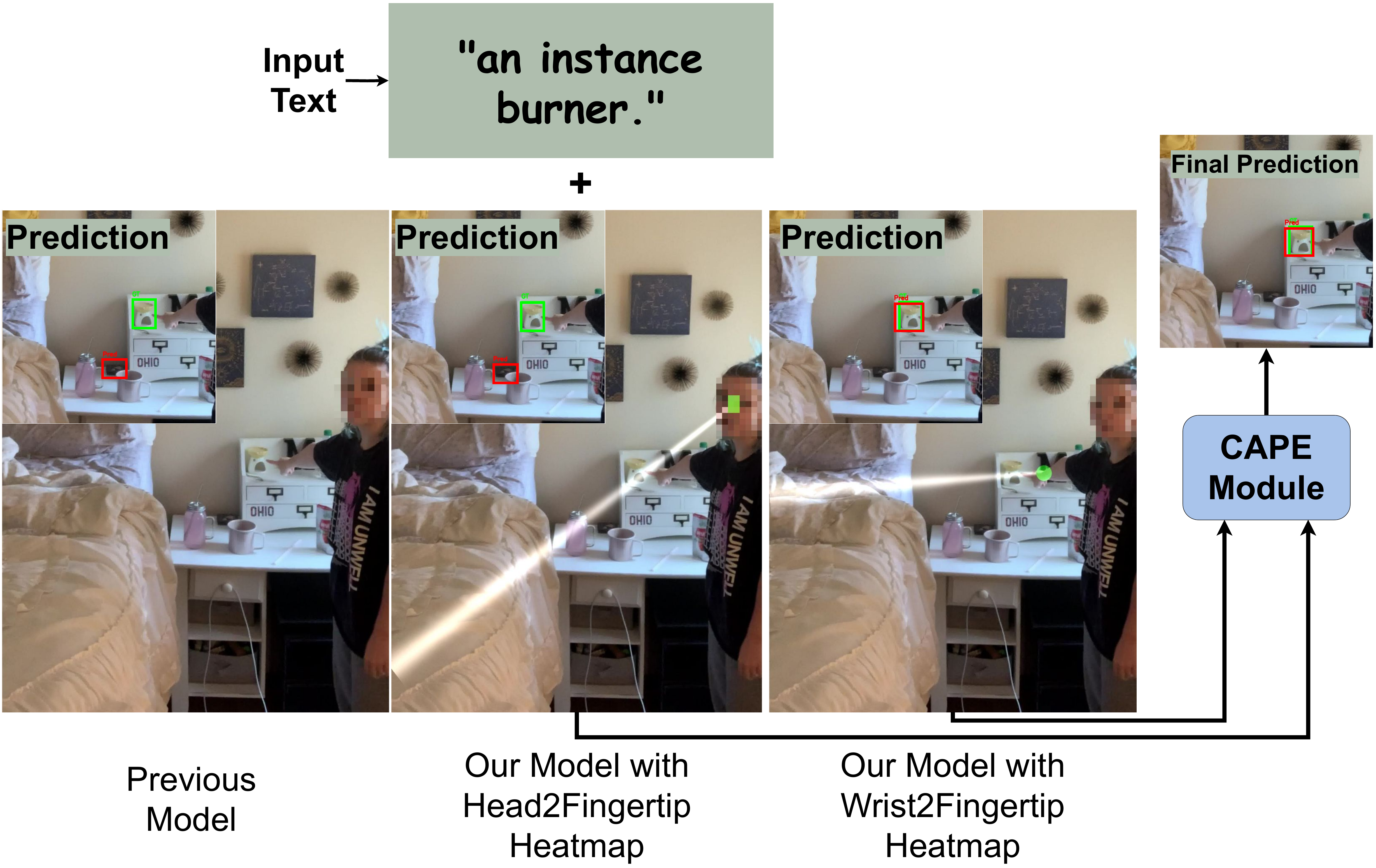}
    \caption{An example where previous models fail. Our $M_{H}$ model also fails due to distractions from other objects along the head-to-fingertip pointing line. In contrast, our $M_{W}$ model correctly identifies the target object, and CAPE selects this as the final prediction. This highlights the flexibility and robustness of our approach.}
    \label{fig:task_description}
\end{figure}

While grounding models~\cite{liu2024grounding} and LMMs~\cite{bai2023qwen,bai2025qwen2,beyer2024paligemma,steiner2024paligemma} have made significant progress in detecting objects mentioned in natural language, they often fall short in ERU task, particularly in ambiguous scenes. 
When multiple instances of the same object type are present, these models tend to detect all matching candidates without the ability to disambiguate and identify the specific target intended by the user. 
Moreover, when the textual instruction itself is vague or ambiguous, grounding models struggle even further, often failing to identify the correct target object or producing no confident prediction at all due to the inherent ambiguity.
This limitation highlights the need for additional disambiguation cues that can help resolve referential ambiguity and enable accurate identification of the intended object.
Therefore, both embodied gesture signals (pointing) and language reference are crucial to identify the referent, aligning with early multimodal interaction studies~\cite{yang1998visual,suhm1999model}.

The ERU task was first defined in \cite{chen2021yourefit} with the YouRefIt dataset, and a method combining textual and visual inputs with PAF~\cite{cao2017realtime} and predicted saliency maps~\cite{kroner2020contextual} was proposed. 
While effective, the model often fails to follow pointing instructions accurately. 
The Touch-Line Transformer~\cite{li2023understanding} addresses this by predicting a Virtual Touch-Line along the head-to-fingertip direction. 
However, it still faces limitations as the pointing line is not explicitly provided as input, despite its success~\cite{zhao2020learning}, and the head-to-fingertip direction does not always align with the target. 
In some cases, accurate pointing is better represented from the hand alone (e.g., wrist-to-fingertip, \cref{fig:task_description}), highlighting the need for more reliable pointing guidance.

To tackle these challenges, we propose a dual-model framework that leverages complementary pointing cues. 
Both models take as input a referring expression and an image with a pointing gesture, along with an additional heatmap that encodes the pointing direction. 
One model utilizes a head-to-fingertip heatmap, while the other uses a wrist-to-fingertip heatmap, capturing different aspects of the pointing behavior. 
Since these two models provide complementary information depending on the scenario, we introduce the CLIP-Aware Pointing Ensemble (CAPE) module to effectively combine the strengths of both models.
In this module, we leverage the CLIP model to compute semantic similarity scores between the input text and candidate object images. 
We use CLIP because it was trained on a large-scale image-text dataset, learning a joint embedding space in which semantically matching images and text are closely aligned.
Additionally, we introduce an auxiliary object center prediction head, providing a supervisory signal that guides the model to more accurately localize the pointed object.

Our contributions are: 
(1) We propose using a pointing heatmap as an additional modality to guide our model to focus more effectively on pointing cues. Specifically, we use detected head, wrist, and fingertip points to construct a Gaussian Ray Heatmap that highlights the approximate pointing area. The embedding of this pointing heatmap is then extracted via a heatmap encoder and provided to the model.
(2) We propose two parallel complementary models to focus on head-to-fingertip and wrist-to-fingertip pointing lines, to individually alleviate inherent challenges of the task.
(3) We present CAPE module to effectively ensemble two complementary models as ensembling has shown strong performance in other fields~\cite{ganaie2022ensemble,zhang2022review,hoang2023fly,yu2024crema,zhang2025delving}.
(4) We introduce an object center prediction head to improve prediction accuracy by emphasizing object localization without requiring precise bounding boxes. 

%% file: sec/2_related_work.tex
\section{Related Work}
\label{sec:relatedWork}

\paragraph{Referring Expression Comprehension (REC)}
\label{subsection2_1}
In REC~\cite{he2023grec,kazemzadeh2014referitgame,nagaraja2016modeling,pi2024perceptiongpt,qiao2020referring,yu2018mattnet}, the goal is to identify a specific region in an image based on a given referring expression. 
In contrast to traditional object detection, REC interprets free-form text and can locate objects from any category, including previously unseen ones. 
Similarly, OV-DETR~\cite{zareian2021open} integrates image and text embeddings from a CLIP model as queries within the DETR~\cite{carion2020end} to generate category-specific bounding boxes. 
ViLD~\cite{gu2021open} distills knowledge from a CLIP teacher model into an R-CNN-like detector~\cite{girshick2014rich}, enabling region embeddings to capture semantic information from language. 
GLIP~\cite{gao2024clip} formulates object detection as a grounding task, using additional grounding datasets to align visual regions with textual phrases. 
DetCLIP~\cite{yao2022detclip} enriches its knowledge base using generated pseudo labels.
While YOLO-World~\cite{cheng2024yolo} extends traditional YOLO object detection~\cite{redmon2016you} to an open-world setting, GroundingDINO~\cite{liu2024grounding} aligns visual and textual features to detect arbitrary objects described by free-form text.
Recently, LMMs~\cite{beyer2024paligemma,abdin2024phi,bai2023qwen} get attention for their superior performance in visual grounding~\cite{chen2025revisiting}.

\vspace{-4mm}

\paragraph{Nonverbal Communication for Referent}
\label{subsection2_2}
Some studies~\cite{stiefelhagen1999gaze,stiefelhagen1999modeling,stiefelhagen2001estimating,stiefelhagen2004natural,yang2007gesture,burger2012two,xiao2014human,holzapfel2004implementation,pateraki2014visual,tanada2024pointing,johari2021gaze,smith2008tracking,valenti2011combining,tonini2023object,gupta2022modular,tafasca2024sharingan,yang2024gaze,tafasca2023childplay,tonini2022multimodal,tafasca2024toward,gupta2024exploring,ryan2025gaze} use gaze target detection as a nonverbal cue, however, gaze alone is unreliable due to distractions and the absence of a clear pointing moment. 
It is typically localized frame by frame and often used to support conversation~\cite{bazzani2013social,capozzi2019tracking}. 
In~\cite{oyama2023exophora}, a multimodal exophora resolution method is proposed to disambiguate demonstrative expressions like “that one” by integrating object categories, pointing gestures, and prior environmental knowledge. 
\cite{constantin2022interactive} introduced an interactive robot dialogue system that uses multimodal interaction and pointing line estimation to accurately identify referent with an iterative correction process.  
Their follow-up work~\cite{constantin2023multimodal} extended this by recognizing unseen objects using a Region Proposal Network and VL-T5 multimodal network~\cite{cho2021unifying}, moving beyond general object classes.
On the other hand, several datasets have been collected to incorporate nonverbal cues, but most remain in simulated domains due to accessibility and controllability~\cite{islam2022caesar,islam2023eqa,alalyani2024embodied,jain2024gesnavi}. 

\vspace{-4mm}

\begin{figure*}[ht!]
    \centering
    \includegraphics[width=1\linewidth]{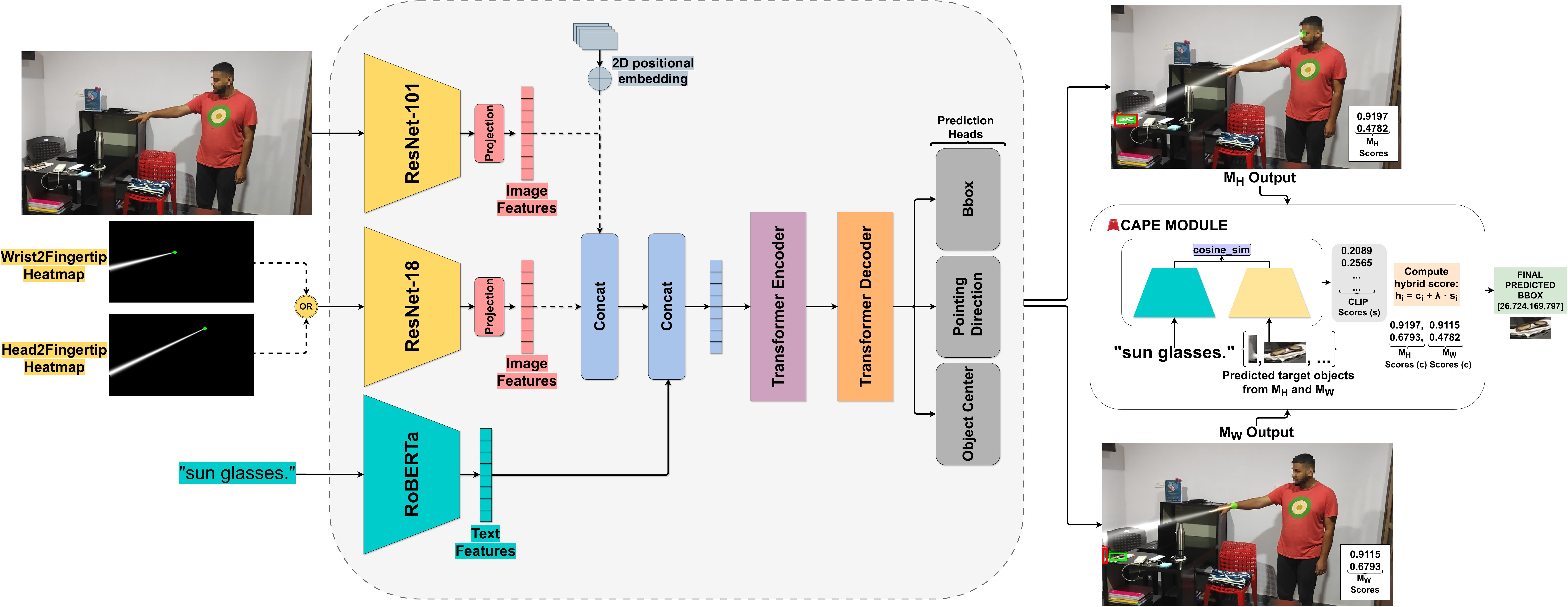}
    \caption{Overview of our approach. $M_H$ and $M_W$ share the same architectural design; therefore, only one model is illustrated. For visualization, we show the two different heatmaps: $M_H$ uses a head-to-fingertip heatmap, while $M_W$ uses a wrist-to-fingertip heatmap.}
    \label{fig:model}
\end{figure*}

\paragraph{Embodied Reference Understanding (ERU)}
\label{subsection2_3}
ERU is a recent advancement of REC that considers the subject’s position while pointing at a reference object. 
Chen et al.~\cite{chen2021yourefit} introduced the task and benchmarks using both verbal (text) and nonverbal (pointing gesture) cues. 
Their model leverages predicted saliency maps and PAF~\cite{cao2017realtime} as gestural features to better perceive pointing direction and detect referents. 
To address ambiguities caused by camera perspective, \cite{shi2022spatial} mapped scenes to 3D coordinates using depth estimation and subject position for spatial attention, but this only yielded modest improvements.  
Li et al.~\cite{li2023understanding} further improved ERU using an MDETR-based~\cite{kamath2021mdetr} model, introducing a “virtual-touch-line” from eye to fingertip and predicting its vector alongside bounding boxes.
In this work, we also use a transformer-based multimodal object detector similar to MDETR~\cite{kamath2021mdetr}. 
However, instead of predicting a single line vector, we extend the visual-touch-line concept to a pointing heatmap representing the focus of attention. 
To capture different pointing variations, we use two different heatmaps processed in parallel networks.
Some recent works \cite{lu2024scaneru,mane2025ges3vig} have extended ERU task to 3D embodied settings. ScanERU~\cite{lu2024scaneru} introduced the first 3D-ERU benchmark by inserting human avatars into existing 3D datasets, but relied on manual placement. 
In contrast, Ges3ViG~\cite{mane2025ges3vig} improves realism and scalability by automating avatar insertion, generating gesture-aware instructions, and incorporating human localization into the grounding pipeline.

%% file: sec/3_method.tex
\section{Methodology}
\label{sec:method}
In ERU, given an RGB image $x_{img} \in \mathbb{R}^{3 \times H \times W} $ and a text input $x_{text} \in \mathbb{N}^L$, we predict a bounding box $x_{bbox} \in \mathbb{R}^4$ corresponding to an object referenced by the text and indicated by a pointing gesture in the image.
We propose using an additional pointing heatmap $x_{phm} \in \mathbb{R}^{1 \times H \times W}$ to supervise the model better focus on the visual pointing reference.

\subsection{Model Architecture}
\label{subsection3_1}

\cref{fig:model} illustrates our overall approach, which consists of two parallel models ($M_H$ and $M_W$; only one is shown since the sole difference is the heatmap encoder input). 
After processing the text, image, and heatmap inputs, the outputs of both models are passed to the CLIP-Aware Pointing Ensemble (CAPE) module for the final decision. 
CAPE combines the outputs of these two complementary models to enhance overall performance.
Specifically, $M_H$ and $M_W$ each comprise a large-scale text encoder, image encoder, and heatmap encoder. 
The outputs of these encoders are concatenated and fed into a transformer~\cite{vaswani2017attention} encoder block. 
A subsequent transformer decoder then processes the encoder output to generate predictions via task-specific heads.

\vspace{-4mm}

\paragraph{Encoders}
We use a pretrained ROBERTA~\cite{liu2019roberta} which is a robustly optimized version of BERT~\cite{devlin2019bert} as the text encoder to obtain textual embeddings from the input text.
To obtain image embeddings, $F_I \in \mathbb{R}^{2048 \times 8 \times 8}$, we utilize a pretrained ResNet-101~\cite{he2016deep}. 
Since the pointing heatmap ($x_{phm}$) is a sparse representation, we choose a lightweight encoder, ResNet-18~\cite{he2016deep}, to embed it: $F_{phm} \in \mathbb{R}^{256 \times 8 \times 8}$
This choice not only improves the computational efficiency of the network but also helps prevent redundant information in the final concatenated feature representation, which could otherwise harm the model's learning capacity.

\vspace{-4mm}

\paragraph{Transformer encoder and decoder}
After obtaining the embeddings from three encoders, we first apply a $1\times1$ convolution to each embedding individually to project their channel dimensions to $256$.
Next, we flatten the spatial dimensions of the embedding into a single dimension, converting them into sequences of tokens. 
For example, for the heatmap embedding: $F^S_{phm} = F(C_{1\times1}(F_{phm})) \in \mathbb{R}^{64 \times 256}$, where $F$ is flatten and $C$ is convolution.
We then concatenate the three embeddings at the sequence level to form the final representation, which is fed into the transformer encoder. 
It processes this concatenated input to learn multimodal representations. 
We feed the multimodal representation output from the transformer encoder into the transformer decoder.
Additionally, we provide a set of learnable object queries and gestural keypoint queries.
The transformer decoder is responsible for generating object output embeddings and gestural output embeddings.

\vspace{-4mm}

\paragraph{Prediction head}
The object and gestural output embeddings produced by the transformer decoder serve as inputs to our prediction heads.
These heads are responsible for predicting candidate bounding boxes for the referent, gestural keypoints (pointing direction prediction head in \cref{fig:model}), which predicts eye, fingertip, wrist coordinates as well as arm class, and center points of the referent.
Finally, we select one bounding box prediction, one center point and a pair of gestural keypoints with the highest confidence scores as the final prediction.
Please note that we use multi-layer perceptrons (MLPs) as the prediction heads.

\subsection{Pointing Heatmap Modality Learning}
Incorporating a heatmap that represents the pointing direction provides a valuable spatial priors for the network, especially considering the fact that the existing models in this field tend to give less attention to the visual pointing.
This explicit encoding pointing direction helps localize the target region by guiding the model's attention toward direction and areas that are more likely to be referenced.
This is particularly beneficial in real-world, cluttered scenes where gesture interpretation is inherently ambiguous.

\vspace{-4mm}

\paragraph{How should we create heatmaps?}
While prior work such as VTL~\cite{li2023understanding} shows that head–to–fingertip cues can often capture the intended pointing direction, our goal is to design a heatmap representation that remains reliable across a wider range of real-world conditions. 
Human pointing gestures are guided by visual attention, and the head–to–fingertip line naturally encodes this gaze–gesture alignment. 
However, relying on this cue alone can be problematic when the performer is looking elsewhere (e.g., toward a robot or camera) or when the head pose becomes ambiguous due to extreme angles or occlusions.
Moreover, close proximity to the object or incomplete arm extension can also break alignment. 
To address these limitations, our approach incorporates both the head–to–fingertip and wrist–to–fingertip lines. 
The wrist–to–fingertip line provides a complementary, locally grounded cue that remains informative even when head orientation is unreliable. 
By combining these two directional sources, we aim to obtain a more robust and consistent representation of pointing intent. Based on this hypothesis, we generate Gaussian Ray Heatmaps for both cues rather than relying solely on explicit line overlap.
Both heatmaps extend a ray toward the image boundary through the fingertip, originating at the eyes for $x^{H}_{phm}$ and at the wrist for $x^{W}_{phm}$.
These heatmaps serve as complementary spatial supervision signals generated through our heatmap process (see \cref{app:heatmap_generation}), capturing not only overlap with a single pointing line but also a broader region that reflects the performer’s focus of attention.

\vspace{-4mm}

\paragraph{How should we integrate these heatmaps?}
We investigate two strategies for integrating the two heatmaps.
The first merges them directly ($x^{H}_{phm} + x^{W}_{phm}$) and feeds the combined input into a single network, testing whether the model can implicitly resolve the two cues and infer the correct pointing direction from pose context.
The second trains two identical networks separately, one with the head-to-fingertip heatmap ($M_{H}$) and one with the wrist-to-fingertip heatmap ($M_{W}$), and ensembles their predictions.
Empirically, the second strategy works substantially better. 
It yields two complementary models, with $M_{H}$ providing the strongest gains. 
In contrast, the merged-heatmap approach offers only marginal improvement. 
The model behaves similarly to the head-only variant, and the extra wrist-to-fingertip signal introduces conflicting directional cues that ultimately reduce accuracy compared to using the head-to-fingertip heatmap alone (see \cref{tab:ablation_general}, Setup \texttt{G}).
We find that the most effective way to leverage heatmaps is to process them with a dedicated encoder whose embeddings are concatenated with the image encoder’s output. This design preserves the pretrained visual backbone while allowing the heatmap encoder to capture richer, modality-specific spatial cues.

\subsection{CLIP-Aware Pointing Ensemble (CAPE)}
We introduce an inference time ensemble module to increase the performance by benefiting from two complementary models, $M_{H}$ and $M_{W}$, since we find that they demonstrate strength in different scenarios.
CAPE is an adaptive scoring method that combines model confidence and CLIP-based similarity in a size-aware manner effectively.
We choose the CLIP model because it was trained to understand and evaluate image-text semantic similarity, which aligns well with our task.
(1) For each prediction (top-2 predictions from both models), we compute a normalized CLIP score~\cite{radford2021learning}. 
We sum pointing models' confidence scores with the CLIP scores to obtain the fused score (CLIP Fusion in \cref{tab:ablation_ensemble}) since the models’ confidence scores also provide reliable insights.
(2) We calculate CLIP scores for the top-1 predictions of both models. 
If the confidence scores of the second-highest predictions exceed a threshold, we also compute their CLIP scores. 
Finally, we select the prediction with the highest CLIP score from among these candidate boxes (CLIP-Only Top-2 + Threshold=0.95 in \cref{tab:ablation_ensemble}).
CAPE applies strategy (1) when the referent is a small object (defined as occupying less than 0.48\% of the image area, following \cite{chen2021yourefit}) to avoid relying solely on the CLIP prediction, as CLIP becomes less reliable for smaller objects.
For all other objects, strategy (2) is used (see \cref{app:cape} for details). 
This hybrid approach leverages the strengths of both signals while adapting to object scale. 
All thresholds and selection rules were tuned on the validation set to avoid test-time overfitting.

\subsection{Gestural Signal Learning}

\paragraph{Referent alignment loss} 
Following \cite{li2023understanding}, we incorporate a referent alignment loss ($L_{RA}$) to enforce consistency between the predicted referent and the VTL. 
The core idea is that a correct referent should be geometrically aligned with the pointing direction.
To quantify this alignment, we compute the cosine similarity between the eye-to-fingertip vector and the eye-to-object vector, defined as:
\begin{equation}
    CS_{p} = \omega((x_f - x_e, y_f - y_e), (x_o - x_e, y_o - y_e))
    \label{eq:cos_sim_pred}
\end{equation}
where ($x_e, y_e$) and ($x_f, y_f$) are the eye and the fingertip coordinates. ($x_o, y_o$) are the center of the predicted bounding box.
$\omega$ is cosine similarity.
We compute the same similarity using GT bounding box to obtain $CS_{GT}$.
The referent alignment loss then penalizes deviations between the predicted and GT alignment:
\begin{equation}
    L_{RA} = ReLU(CS_{p} - CS_{GT})
\end{equation}
Please note that for both $CS_{p}$ and $CS_{GT}$, we use GT eye and fingertip coordinates to ensure accurate directional representation.
We apply this loss during the training of the model $M_{H}$, which uses the head-to-fingertip heatmap as reference.
For training the $M_{W}$ model, which utilizes the wrist-to-fingertip heatmap, we replace the eye coordinates with wrist coordinates.
The rest remains unchanged, as we similarly aim to maximize the directional correlation.

\vspace{-4mm}

\paragraph{Object center prediction}
In our task, identifying the correct referent and accurately localizing it are two distinct but equally important challenges.
To enhance the localization capability of the model, we decouple the prediction of the object center from the rest of the bounding box regression.
Specifically, we introduce an additional prediction head implemented as a MLP at the end of the network. 
This head is dedicated to predicting the ($x, y$) coordinates of the referent object's center, independently from the standard bounding box regression head.
During training, we supervise this center prediction using the L1 loss between the predicted and GT center coordinates:
\begin{equation}
    L_{center} = ||(x^{GT}_o, y^{GT}_o) - (x^p_o, y^p_o)||_1
\end{equation}
This explicit supervision encourages the model to focus on spatial alignment and leads to more accurate object localization, especially in scenes with dense or overlapping objects.
Moreover, this center prediction serves as an auxiliary task, which improves the representational capacity of the model by encouraging the visual backbone to learn richer geometric and spatial features. 
We apply this auxiliary head and its corresponding loss to both models. 
Since this head is used only during training, it introduces no additional computational overhead during inference.

\vspace{-4mm}

\paragraph{Gesture prediction}
In addition to the bounding box and object center predictions, our model incorporates several auxiliary heads to predict task-specific features, including eye (wrist in $M_{W}$) and fingertip coordinates, and arm classification (See prediction head in \cref{subsection3_1}). 
For each prediction, we apply a standard L1 loss between the predicted and GT data, which encourages spatial attention to task-relevant regions such as the eyes and fingertips, enhancing the model’s understanding of referential gestures.
Moreover, we apply cross-entropy loss to classify whether the predicted gestural keypoints are correct.

\vspace{-4mm}

\paragraph{Total loss}
We define our objective as follows:
\begin{equation}
    L = \lambda_1 L_{b} + \lambda_2 L_{RA} + \lambda_3 L_{center} + \lambda_4 L_{g} + \lambda_5 L_{t} + \lambda_6 L_{c}
\end{equation}
where $L_{b}$ is bounding box loss and it is the combination of L1 and GIoU losses as in DETR-like methods~\cite{carion2020end,li2022dn,meng2021conditional,zhang2022dino,zhu2020deformable,liu2024grounding}.
$L_{RA}$ denotes referent alignment loss, $L_{center}$ indicates object center prediction loss, and $L_{g}$ is gesture prediction loss.
$L_{t}$ and $L_{c}$ are soft token loss and contrastive loss respectively to help visual and textural signals alignment as in \cite{kamath2021mdetr}.
We empirically determine the optimal coefficients on the validation set (See \cref{app:hyperparameters}).

%% file: sec/4_experiments.tex
\begin{table*}[t]
    \centering
    \footnotesize
    \begin{tabular}{l|cccc|cccc|cccc}
        \toprule
        IoU Threshold for mAP & \multicolumn{4}{c|}{0.25} & \multicolumn{4}{c|}{0.50} & \multicolumn{4}{c}{0.75} \\
        \midrule
        Object Sizes & \multicolumn{1}{c}{All} & \multicolumn{1}{c}{S} & \multicolumn{1}{c}{M} & \multicolumn{1}{c}{L} & \multicolumn{1}{|c}{All} & \multicolumn{1}{c}{S} & \multicolumn{1}{c}{M} & \multicolumn{1}{c}{L} & \multicolumn{1}{|c}{All} & \multicolumn{1}{c}{S} & \multicolumn{1}{c}{M} & \multicolumn{1}{c}{L} \\
        \midrule
        PaliGemma2~\cite{steiner2024paligemma}    & 58.8 & 29.0 & 53.5 & 75.8 & 46.9 & 22.1 & 50.8 & 68.0 & 31.7 & 6.2 & 34.1 & 54.8 \\
        Qwen2.5vl~\cite{bai2025qwen2}  & 38.9 & 17.0 & 41.8 & 58.0 & 31.0 & 11.1 & 33.6 & 48.1 & 20.0 & 5.7 & 19.8 & 34.5 \\
        Grounding DINO~\cite{liu2024grounding} & 57.9 & 38.0 & 60.9 & 74.9 & 54.9 & 35.7 & 59.3 & 69.6 & \textbf{42.3} & \textbf{22.7} & \textbf{45.9} & \textbf{58.4} \\
        \midrule
        FAOA \cite{yang2019fast} & 44.5 & 30.6 & 48.6 & 54.1 & 30.4 & 15.8 & 36.5 & 39.3 & 8.5 & 1.4 & 9.6 & 14.4 \\
        ReSC \cite{yang2020improving}  & 49.2 & 32.3 & 54.7 & 60.1 & 34.9 & 14.1 & 42.5 & 47.7 & 10.5 & 0.2 & 10.6 & 20.1 \\
        YourRefit PAF \cite{chen2021yourefit} & 52.6 & 35.9 & 60.5 & 61.4 & 37.6 & 14.6 & 49.1 & 49.1 & 12.7 & 1.0 & 16.5 & 20.5 \\
        YourRefit Full \cite{chen2021yourefit} & 54.7 & 38.5 & 64.1 & 61.6 & 40.5 & 16.3 & 54.4 & 51.1 & 14.0 & 1.2 & 17.2 & 23.3 \\
        REP \cite{shi2022spatial} & 58.8 & 44.7 & 68.9 & 63.2 & 45.7 & 25.4 & 57.7 & 54.3 & 18.8 & 3.8 & 22.2 & 29.9 \\
        Touch-Line-EWL \cite{li2023understanding} & {69.5} & \underline{56.6} & {71.7} & 80.0 & 60.7 & {44.4} & 66.2 & 71.2 & 35.5 & {11.8} & {38.9} & 55.0 \\
        Touch-Line-VTL \cite{li2023understanding} & \underline{71.1} & 55.9 & \underline{75.5} & \underline{81.7} & \underline{63.5} & \underline{47.0} & \underline{70.2} & \textbf{73.1} & \underline{39.0} & \underline{13.4} & \underline{45.2} & \underline{57.8} \\
        \midrule
        Ours (CAPE) & \textbf{75.0} & \textbf{63.2} & \textbf{80.2} & \textbf{81.8} & \textbf{65.4} & \textbf{49.5} & \textbf{74.3} & \underline{72.7} & {35.7} & \underline{13.4} & {40.1} & {53.5} \\
        \bottomrule
    \end{tabular}
    \caption{Comparison of our model with prior works, SOTA LMMs, and Grounding-DINO in terms of mean Average Precision (mAP) at different IoU thresholds, across various object sizes, on the YouRefIt dataset~\cite{chen2021yourefit}.}
    \label{tab:comparison_iou}
\end{table*}

\section{Experimental Results}
\label{sec:experiments}

\begin{figure*}
    \centering
    \includegraphics[width=0.8\linewidth]{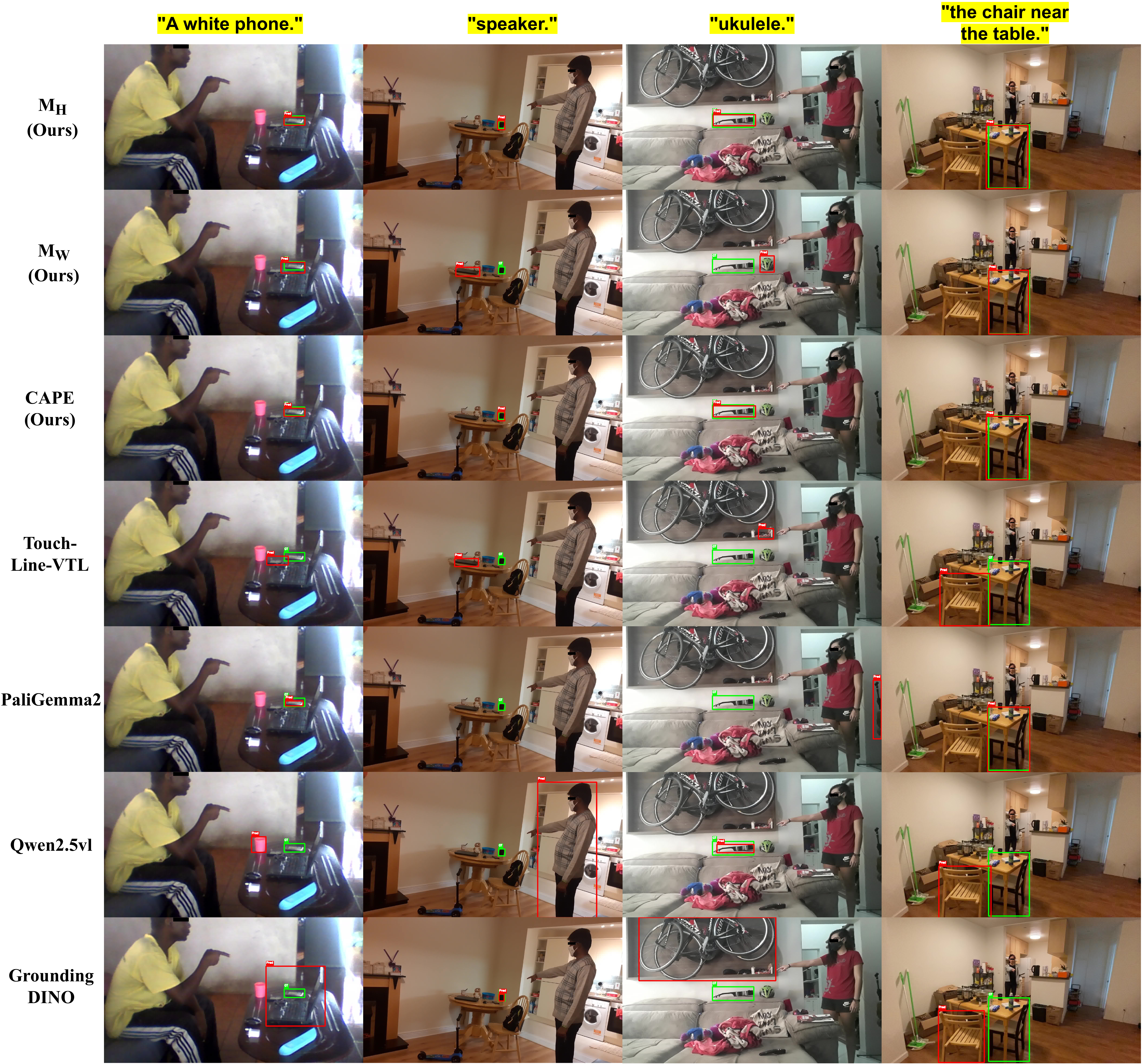}
    \caption{Qualitative comparison of our models with SOTA Touch-Line Transformer~\cite{li2023understanding}.}
    \label{fig:comparison}
\end{figure*}

\paragraph{Dataset}
For both training and testing, we use the YouRefIt dataset \cite{chen2021yourefit}, which contains 2,950 training and 1,245 test images. 
Text instructions and the annotations of bounding boxes, as well as pointing line coordinates, are provided by \cite{chen2021yourefit, li2023understanding}.
Additionally, we evaluate our approach on the unseen and more challenging ISL Pointing dataset~\cite{constantin2022interactive}, and a simulation dataset named CAESAR~\cite{islam2022caesar} (for results see appx. \cref{app:tab:comparison_iou_CAESAR}) to assess its generalization capability.

\vspace{-4mm}

\paragraph{Evaluation}
For fair evaluation and comparison, we follow prior work~\cite{chen2021yourefit} and report mean Average Precision (mAP) under three Intersection-over-Union (IoU) thresholds: 0.25, 0.50, and 0.75.
Additionally, mAP scores are reported with respect to object size, categorized as Small (S), Medium (M), and Large (L) which based on the ratio of the object area to the image area using two thresholds: $0.48\%$ and $1.76\%$.
In addition to this standard evaluation protocol, we introduce two metrics for assessing referential grounding performance: CLIP score and center coordinate distance ($C_D$).
To compute the CLIP score, we extract features from the text input and the cropped region corresponding to the predicted bounding box using CLIP's text and image encoders. 
We then calculate the cosine similarity between these feature vectors to quantify the semantic alignment between the predicted object and the text.
For $C_D$, we compute the L1 distance between the center coordinates of the predicted and the GT bounding boxes, providing a direct measure of spatial alignment.

\vspace{-4mm}

\paragraph{Implementation}
We use the AMSGrad optimizer~\cite{reddi2019convergence,kingma2014adam} during training and train our models for 30 epochs.
Both the transformer encoder and decoder consist of $6$ layers, each with $8$ attention heads and an MLP dimension of $2048$.
We apply dropout~\cite{srivastava2014dropout} with $p=0.1$ in every layer of both the transformer encoder and decoder.
The learning rate is set to $1e-4$ for the text encoder and $5e-5$ for the remaining components.
All experiments are conducted on a single NVIDIA RTX A6000 GPU with a batch size of 4.

\begin{table}[t]
    \centering
    %\tiny
    \setlength{\tabcolsep}{4pt}
    \resizebox{\linewidth}{!}{\begin{tabular}{lcccc|cccc}
        \toprule
        & \multicolumn{4}{c}{CLIP Score $\uparrow$ } & \multicolumn{4}{c}{$C_D$ $\downarrow$ } \\
        \midrule
        Objet Sizes & \multicolumn{1}{c}{All} & \multicolumn{1}{c}{S} & \multicolumn{1}{c}{M} & \multicolumn{1}{c|}{L} & \multicolumn{1}{c}{All} & \multicolumn{1}{c}{S} & \multicolumn{1}{c}{M} & \multicolumn{1}{c}{L} \\
        \midrule
        PaliGemma2     & 0.2488 & 0.2302 & 0.2446 & 0.2659 & 0.4684 & 0.5784 & 0.4307 & 0.4175 \\
        Qwen2.5vl-3b   & 0.2408 & 0.2261 & 0.2388 & 0.2568 & 0.7288 & 0.7074 & 0.6893 & 0.7891 \\
        Grounding DINO & 0.2437 & 0.2267 & 0.2413 & 0.2621 & 0.3496 & 0.2946 & 0.4521 & 0.3104 \\
        Touch-Line-EWL & 0.2456 & 0.2312 & 0.2435 & 0.2615 & 0.3168 & 0.3006 & 0.2903 & 0.3564 \\
        Touch-Line-VTL & 0.2456 & 0.2308 & 0.2440 & 0.2615 & 0.2843 & 0.2809 & 0.2276 & 0.3393 \\
        \midrule
        Ours (CAPE) & \textbf{0.2661} & \textbf{0.2642} & \textbf{0.2671} & \textbf{0.2670} & \textbf{0.2476} & \textbf{0.2137} & \textbf{0.2241} & \textbf{0.3023} \\
        \bottomrule
    \end{tabular}}
    \caption{Clip and $C_D$ scores on the YouRefIt dataset.}
    \label{tab:comparison_new_metrics}
\end{table}

\begin{table}[]
    \centering
    \tiny
    \setlength{\tabcolsep}{5pt}
    \resizebox{\linewidth}{!}{\begin{tabular}{l|ccccc}
        \hline
        Method & IoU=0.25 & IoU=0.5 & IoU=0.75 & CLIP $\uparrow$ & $C_D$ $\downarrow$ \\
        \hline
        PaliGemma2     & 47.2 & 39.5 & \textbf{31.6} & 0.2465 & 0.7449 \\
        Qwen2.5vl-3b   & 32.5 & 32.1 & 29.2 & 0.2264 & 0.8418 \\
        Grounding DINO & 27.5 & 27.5 & 26.3 & 0.2082 & 0.7956 \\
        Touch-Line-EWL & 45.0 & 35.8 & 22.0 & 0.2436 & 0.5160 \\
        Touch-Line-VTL & 47.7 & 36.7 & 17.4 & 0.2473 & 0.5147 \\
        \hline
        Ours, $M_W$  & 48.2 & 33.6 & 20.0 & 0.2453 & 0.4735 \\
        Ours, $M_H$  & 42.7 & 30.0 & 17.3 & 0.2449 & 0.6795 \\
        Ours, CAPE & \textbf{54.5} & \textbf{42.7} & {25.5} & \textbf{0.2642} & \textbf{0.4403} \\
        \hline
    \end{tabular}}
    \caption{Quantitative results on unseen ISL pointing dataset~\cite{constantin2022interactive}.}
    \label{tab:quantitative_results_ISL}
\end{table}

\subsection{Results}
\cref{tab:comparison_iou} compares the performance of our method with existing approaches on the YouRefIt benchmark.
Our model achieves SOTA results in most cases, despite being trained for only 30 epochs on a single GPU with a batch size of 4.
In contrast, Touch-Line models~\cite{li2023understanding} were trained for 200 epochs with a batch size of 56.
This clearly demonstrates the efficiency and effectiveness of our approach.
However, our optimized training setup affects bounding box precision, contributing to the suboptimal performance at $IoU = 0.75$, alongside limitations related to the simpler backbone. At $IoU = 0.75$, Grounding DINO with a more complex backbone achieves the best performance. 

Further, we evaluate recent LMMs on this task, and they fall significantly short of our model's performance, despite being among the largest models and trained for visual grounding purposes. These results show using complementary pointing heatmap modalities in addition to image and text is essential. 
Besides, despite having two separate models, our approach shows better inference-time performance (2.44 fps) compared to Paligemma2 (1.24fps) and Qwen2.5vl (0.5fps). 
\cref{tab:comparison_new_metrics} presents further comparison using two new metrics in this field: CLIP score and $C_D$. 
Higher CLIP scores indicate that our predicted bounding boxes have stronger semantic alignment with the input text, suggesting better object identification and improved bounding box precision.
Similarly, our method achieves the best performance under the $C_D$ metric. 
This confirms that our model produces more spatially accurate predictions.
In \cref{tab:quantitative_results_ISL}, the experimental results on unseen ISL dataset show that our CAPE model surpasses the SOTA Touch-Line model as well as the evaluated LMMs, except for the IoU=0.75 setup.
This demonstrates the strong generalization capability of our method.
In \cref{fig:comparison}, we present sample images with predicted and GT bounding boxes. 
It is clearly seen from LMMs' predictions, ambiguous text without pointing cues results in poor object localization outside of the pointing area and confusion with similar objects. 
For instance, Paligemma2 in third example and Qwen2.5vl in second and fourth examples detect wrong objects. 
On the other hand, the LMMs are strong at providing precise bounding box predictions due to extensive training and a robust backbone.
Furthermore, even if Touch-Line model use pointing cues since it follows a pointing line instead of heatmaps, this let model to detect other objects in the pointing direction. In contrast, as shown in the examples, CAPE leverages explicit pointing heatmaps, enabling it to more accurately localize the intended referent. Also, it enables to choose final prediction from which of the more accurate model.

\begin{table}[t]
    \centering
    \setlength{\tabcolsep}{5pt}
    \resizebox{\linewidth}{!}{\begin{tabular}{c|lccccc}
        \toprule
        Setup & Method & IoU=0.25 & IoU=0.5 & IoU=0.75 & CLIP $\uparrow$ & $C_D$ $\downarrow$ \\
        \midrule
        \texttt{A} & Baseline & 71.2 & 60.1 & 32.8 & 0.2469 & 0.2662 \\
        \texttt{B} & \texttt{A} + object center prediction & 70.8 & 61.6 & 34.6 & 0.2446 & 0.2707 \\
        \texttt{C} & \texttt{A} + W2F heatmap & 68.9 & 59.7 & 32.4 & 0.2451 & 0.2984 \\
        \texttt{D} & \texttt{A} + H2F heatmap & 71.9 & 62.8 & 33.8 & 0.2458 & 0.2694 \\
        \texttt{E} & \texttt{B} + W2F heatmap ($M_{W}$) & 69.6 & 60.7 & 31.5 & 0.2448 & 0.2770 \\ 
        \texttt{F} & \texttt{B} + H2F heatmap ($M_{H}$) & 72.9 & 62.3 & 35.1 & 0.2457 & 0.2490 \\
        \texttt{G} & \texttt{A} + W2F heatmap + H2F heatmap & 70.2 & 60.2 & 33.8 & 0.2448 & 0.2744 \\
        \texttt{H} & Full model - Ensemble of \texttt{E} and \texttt{F} & \textbf{75.0} & \textbf{65.4} & \textbf{35.7} & \textbf{0.2661} & \textbf{0.2476} \\
        \bottomrule
    \end{tabular}}
    \caption{Ablation study on YouRefIt dataset for our contributions.}
    \label{tab:ablation_general}
\end{table}

\subsection{Analysis}
In this section, we provide a detailed analysis of the optimality of our design choices, including heatmap generation and the ensemble method. 
The baseline in \cref{tab:ablation_general} (Setup \texttt{A}) refers to a straightforward training of our model without incorporating any of the proposed contributions.

\vspace{-4mm}

\paragraph{Object center prediction}
The object center prediction improves mAP at $IoU=0.5$ and $IoU=0.75$ but reduces performance at $IoU=0.25$, as well as CLIP Score and $C_D$.
Notably, when combined with heatmap (Setup \texttt{E} and \texttt{F} in \cref{tab:ablation_general}), it boosts most metrics. 
The object center prediction alone enhances geometric precision more than semantic alignment, while pairing it with spatially dense cues makes it a more effective refining signal.

\begin{table}[h]
    \centering
    %\tiny
    \resizebox{\linewidth}{!}{\begin{tabular}{lccccc}
        \toprule
         & IoU=0.25 & IoU=0.5 & IoU=0.75 & CLIP $\uparrow$ & $C_D$ $\downarrow$ \\
        \midrule
        Channel-wise input & 68.7 & 58.5 & 33.4 & 0.2449 & 0.2959 \\
        Channel-wise feature & 72.6 & 58.4 & 27.9 & 0.2457 & 0.2598 \\
        Embedding feature & \textbf{72.9} & \textbf{62.2} & \textbf{35.1} & \textbf{0.2458} & \textbf{0.2490} \\
        \bottomrule
    \end{tabular}}
    \caption{Ablation of heatmap injection methods using Setup \texttt{F}.}
    \label{tab:ablation_feeding_heatmap}
\end{table}

\vspace{-4mm}

\paragraph{Where to inject heatmap}
We compare several heatmap injection strategies in \cref{tab:ablation_feeding_heatmap}.
The \textit{channel-wise input} method, which feeds the heatmap as a fourth image channel, slightly hurts performance, likely because the heatmap is sparsely represented compared to the image content. 
Inspired by the head location prompting strategy in~\cite{ryan2025gaze}, the \textit{channel-wise feature} approach adds learned heatmap embeddings to the visual tokens, which is lightweight but provides only limited gains.
In contrast, the \textit{embedding feature} setup uses a dedicated CNN to encode the heatmap before fusing it with image and text embeddings, and delivers the strongest improvements across all metrics.
The experiments show that effective heatmap injection requires the model to interpret the heatmap as a structured spatial signal rather than as a raw image channel or a light additive bias on visual tokens.
Besides, the improvements seen in the \textit{embedding feature} method support the gains observed in the main ablations (Setups \texttt{D} and \texttt{F}), where the H2F heatmap consistently strengthens localization and reduces $C_D$.

\begin{table}[t]
    \centering
    \setlength{\tabcolsep}{5pt}
    \resizebox{\linewidth}{!}{\begin{tabular}{lccccc}
        \toprule
        Ensemble method & IoU=0.25 & IoU=0.5 & IoU=0.75 & CLIP $\uparrow$ & $C_D$ $\downarrow$ \\
        \midrule
        Confidence-Only & 73.4 & 64.1 & 34.1 & 0.2578 & 0.2470 \\
        CLIP-Only (Top-1) & 73.5 & 64.2 & 35.4 & 0.2644 & 0.2391 \\ 
        CLIP-Only (Top-2 + Threshold=0.95) & 73.8 & 64.1 & 35.5 & {0.2646} & \textbf{0.2377} \\
        CLIP Fusion & 73.6 & 64.4 & 34.5 & 0.2459 & 0.2422 \\ 
        CAPE & \textbf{75.0} & \textbf{65.4} & \textbf{35.7} & \textbf{0.2661} & 0.2476 \\
        \bottomrule
    \end{tabular}}
    \caption{Ablation of different ensemble methods. See \cref{app:ensemble}.}
    \label{tab:ablation_ensemble}
\end{table}

\vspace{-4mm}

\paragraph{CLIP-Aware Pointing Ensemble (CAPE)}
We evaluate several ensemble strategies for selecting the final referent box from the two pointing models, as shown in \cref{tab:ablation_ensemble}.
(1) \textit{Confidence-Only} selects the Top-1 prediction with the highest model confidence.
(2) \textit{CLIP-Only (Top-1)} chooses the Top-1 box with the highest CLIP similarity.
(3) \textit{CLIP-Only (Top-2 + Threshold$=0.95$)} additionally considers each model’s top-2 predictions when their confidence scores exceed $0.95$.
(4) \textit{CLIP Fusion} combines normalized CLIP and confidence scores from the top-2 predictions of both models and selects the box with the highest total score.
(5) \textit{CAPE} adaptively switches between (3) and (4): it uses (4) for small objects, where CLIP alone is unreliable, and (3) otherwise.
The confidence threshold in (3) and the small-object rule in CAPE were tuned on the validation set to avoid test-time bias.
Across these strategies, the hybrid design in CAPE aligns with observations from the main ablations: the two models provide complementary strengths, CLIP-based scoring refines semantic alignment for most objects, and confidence-based cues are especially important for small or ambiguous targets. 
This contributes to the strong gains seen in the full model (Setup \texttt{H}). 

\vspace{-4mm}

\begin{figure}
    \centering
    \includegraphics[width=1\linewidth]{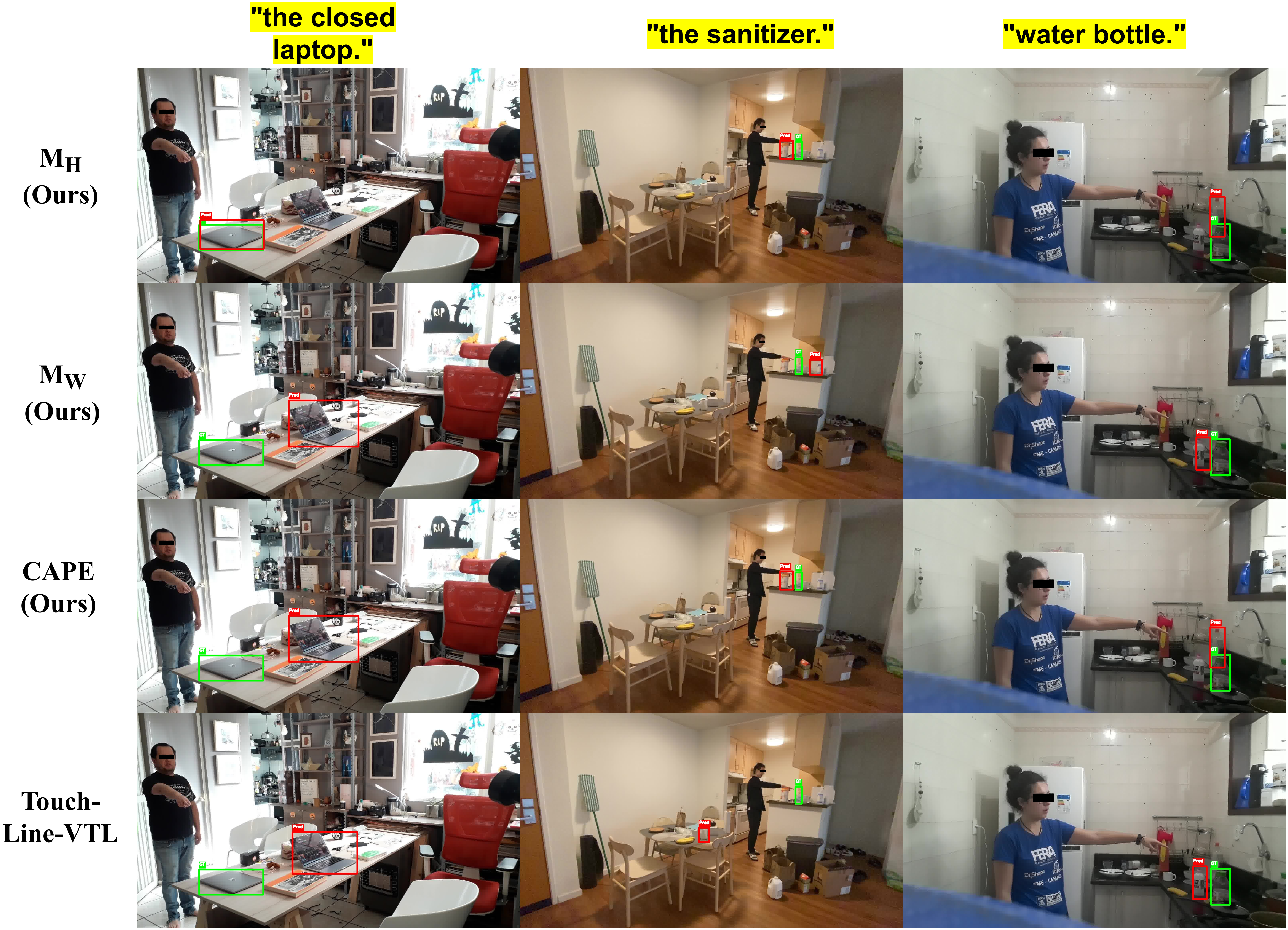}
    \caption{Failure cases from different models. The sample images are from YouRefIt dataset.}
    \label{fig:failures}
\end{figure}

\paragraph{Failures}
In \cref{fig:failures}, we present examples of failure cases.
In the first example (column), although our model $M_{H}$ correctly detects the target object, CAPE selects the prediction from $M_{W}$, which is incorrect in this case.
In the second example, both of our models, as well as the Touch-Line models, produce incorrect predictions.
However, guided by the pointing information, the predictions from $M_{H}$ and $M_{W}$ lie on the pointing line and are very close to the GT object. 
In contrast, the Touch-Line predictions are entirely off the pointing line and significantly distant from the GT object.
In the third column, the bottles are extremely difficult to distinguish, even with explicit pointing cues.
Nevertheless, the predicted bounding boxes cover other bottles adjacent to the GT one, which is acceptable, as they overlap with the pointing line and are semantically consistent with the text.

%% file: sec/5_conclusion.tex
\section{Conclusion}
\label{sec:conclusion}

We address Embodied Reference Understanding by overcoming the limitations of relying on a single pointing line, using our dual-model framework to leverage complementary Gaussian ray heatmaps for richer supervision and improved target detection.
Combined through the CLIP-Aware Pointing Ensemble (CAPE) and enhancing the model with an auxiliary object center prediction, our approach achieves strong improvements on the YouRefIt, ISL pointing, and CAESAR datasets. Our results demonstrate that combining multiple pointing cues and in general, multimodal signals, leads to more accurate and robust referent understanding in complex visual scenes. Beyond empirical gains, our work demonstrates a general principle of multimodal cue integration: fusing complementary spatial and semantic signals resolves ambiguities that single modalities cannot, providing a principle for robust referent understanding in complex visual scenes.

\vspace{-4mm}

\paragraph{Limitations and Future Work} 
The model occasionally selects incorrect objects along the pointing direction due to the absence of depth information, and the CNN backbone limits bounding box precision. Future work could incorporate depth or 3D cues and adopt advanced backbones such as Swin Transformers.
Additionally, formalizing a theoretical framework for multimodal cue integration and exploring adaptive weighting or attention mechanisms could further improve flexibility and robustness, allowing the system to optimally leverage multiple cues across diverse scenarios.

\vspace{-4mm}

\paragraph{Acknowledgment}
This work was supported in part by the European Union’s Horizon research and innovation program, project Meetween (101135798) and project DVPS (Diversibus Viis Plurima Solvo) (101213369), and KIT Campus Transfer GmbH (KCT) under contract from Carnegie-AI LLC.

%\vspace{-4mm}

%% file: sec/X_supp.tex
\setcounter{page}{1}

\maketitlesupplementary

\section{Heatmap Generation}
\label{app:heatmap_generation}
The heatmap generation method is a key component of our approach.
Prior work~\cite{li2023understanding} shows that the referent often aligns with the head-to-fingertip line, which inspired our Gaussian Ray Heatmap generation.
While this generally holds when a person points clearly with an extended arm, it is not always reliable.
Alignment can be disrupted if the person looks elsewhere (e.g., at a robot, camera, or another person), due to camera perspective, or when a person is close to the object and points using wrist motion.
In such cases, the wrist-to-fingertip line aligns more accurately with the referent.
To handle this variability, we generate separate heatmaps for both cues.
\begin{itemize}
    \item $x^{H}_{phm}$: Heatmap from head-to-fingertip.
    \item $x^{W}_{phm}$: Heatmap from wrist-to-fingertip.
\end{itemize}

\subsection{Head-to-Fingertip Heatmap}
To generate this heatmap, we take the head and fingertip as reference points and construct a ray starting at the head and extending toward the image boundary in the fingertip direction.
In the YouRefIt dataset~\cite{chen2021yourefit}, we use the detailed annotations provided in \cite{chen2021yourefit, li2023understanding}, which include eye, elbow, wrist, and fingertip coordinates.
These keypoints allow us to reliably define the reference direction for heatmap generation. 

In real-world settings where such annotations are unavailable, estimating these points is straightforward.
For eye position, face detection followed by using the center of the detected face provides a strong approximation.
Similarly, modern pose estimation models offer accurate wrist and fingertip predictions.
For our experiments on the ISL Pointing and CAESAR datasets, we obtain necessary coordinates using OpenPose~\cite{cao2019openpose}, which delivers robust and reliable keypoint detection.

\subsection{Wrist-to-Fingertip Heatmap}
We follow the same pipeline used for head-to-fingertip heatmap generation, but replace the eye coordinates with wrist coordinates and the fingertip coordinates remain the same.

\subsection{How to Create Heatmap}
\begin{figure*}
    \centering
    \includegraphics[width=1\linewidth]{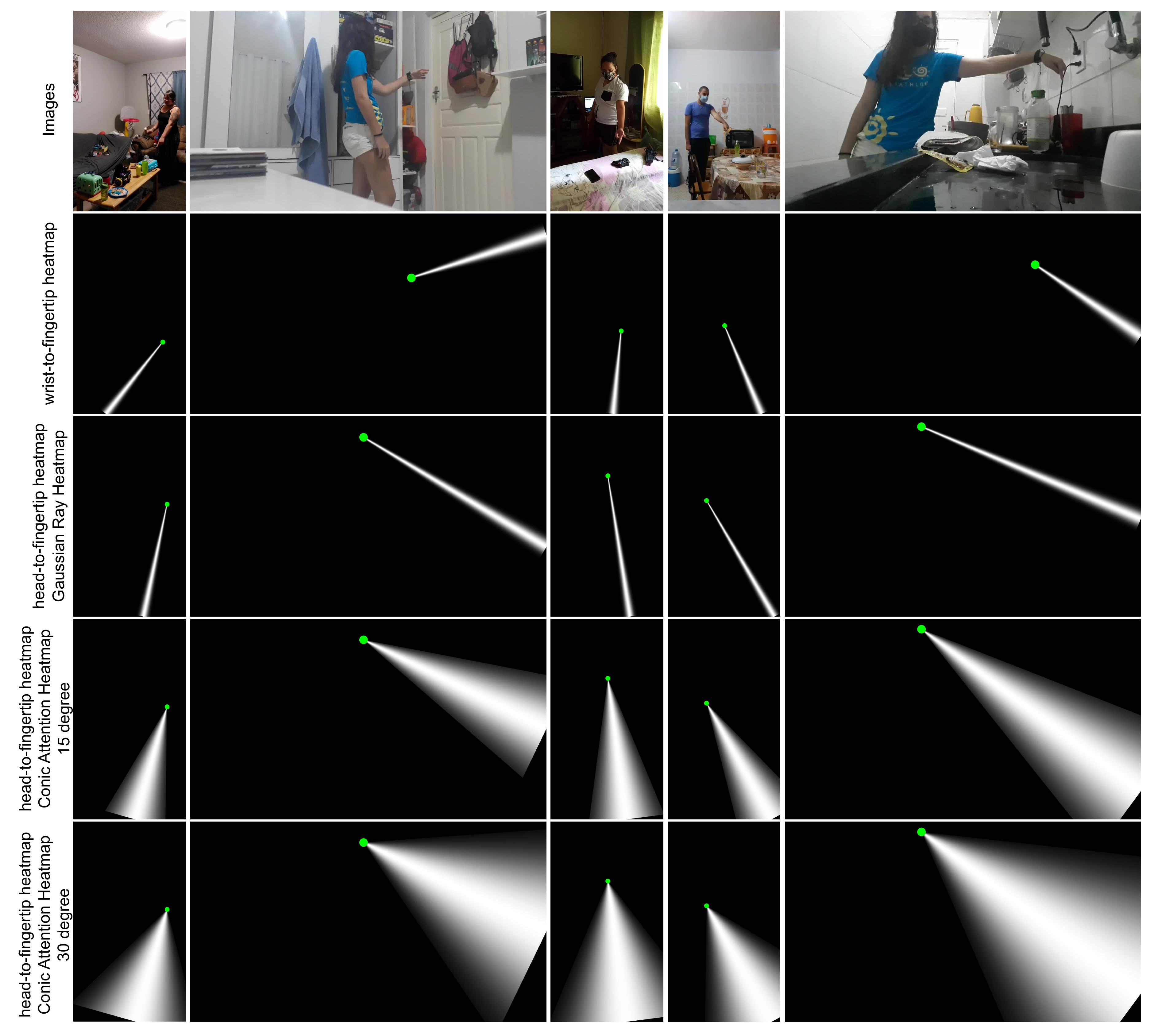}
    \caption{This figure presents various heatmaps. The first row shows the input images. The second row displays wrist-to-fingertip heatmaps used for training $M_{W}$, while the third row contains head-to-fingertip heatmaps used for training $M_{H}$. The fourth and fifth rows show Conic Attention Heatmaps with $15^\circ$ and $30^\circ$ angles, respectively, which are used in the ablation study which we present in \cref{tab:ablation_heatmap_degree}. }
    \label{fig:enter-label}
\end{figure*}

We evaluate the Conic Attention Heatmap using $15^{\circ}$ and $30^{\circ}$ cone angles, as well as the Gaussian Ray Heatmap, which produces more localized activation.
As shown in \cref{tab:ablation_heatmap_degree}, the Gaussian Ray Heatmap achieves the best performance.
This suggests that broader attention regions may be less effective in practice, likely due to the high density of objects in typical scenes (see \cref{fig:enter-label}).
Therefore, larger heatmaps tend to cover multiple distractors, reducing the discriminative value of the spatial signal.

\begin{table}[t]
    \centering
    %\tiny
    \resizebox{\linewidth}{!}{\begin{tabular}{lccccc}
        \toprule
        Heatmap style & IoU=0.25 & IoU=0.5 & IoU=0.75 & CLIP $\uparrow$ & $C_D$ $\downarrow$ \\
        \midrule
        Gaussian Ray Heatmap ($\sigma=25$) & 72.9 & 62.2 & 35.1 & 0.2457 & 0.2490 \\
        Conic Attention Heatmap ($15^{\circ}$) & 70.0 & 60.0 & 33.9 & 0.2464 & 0.2632 \\
        Conic Attention Heatmap ($30^{\circ}$) & 71.5 & 59.4 & 30.8 & 0.2463 & 0.2764 \\
        \bottomrule
    \end{tabular}}
    \caption{Comparison of different heatmap generation approaches.}
    \label{tab:ablation_heatmap_degree}
\end{table}

\section{Ensemble Methods}
\label{app:ensemble}
In this section, we present the details of the ensemble methods we explored.
Each of our models produces $N$ predictions, with each prediction accompanied by a confidence score.
Before performing ensembling, we sort these $N$ predictions in descending order based on their confidence scores.
We denote the sorted prediction list from model $M_{H}$ as $P^N_{H} \in \mathbb{R}^{4 \times N}$, and from model $M_{W}$ as $P^N_{W} \in \mathbb{R}^{4 \times N}$.
Similarly, we denote the confidence score of model $M_{H}$ as $C^N_{H} \in \mathbb{R}^{1 \times N}$, and from model $M_{W}$ as $C^N_{W} \in \mathbb{R}^{1 \times N}$.

\subsection{Confidence-Only}
\label{app:confidence_only}
In the Confidence-Only approach, we focus solely on the top-1 prediction from each model.
We take the bounding box with the highest confidence score from $M_{H}$ and another from $M_{W}$.
These two predictions are then compared, and the one with the higher confidence score is selected as the final prediction.

\begin{algorithm}[H]
\caption{Algorithm of Confidence-Only ensembling method.}
\label{app:alg:confidence_only}
\begin{algorithmic}[1]
\Require Top-1 predictions from models $M_{H}$ and $M_{W}$: $(b^{H}, c^{H})$ and $(b^{W}, c^{W})$
\Ensure Final bounding box prediction $b^*$
\If{$c^{H} \geq c^{W}$}
    \State $b^* \leftarrow b^{H}$
\Else
    \State $b^* \leftarrow b^{W}$
\EndIf \\
\Return $b^*$
\end{algorithmic}
\end{algorithm}

\subsection{CLIP-Only (Top-1)}
\label{app:clip_only_top1}
In this method, similar to the Confidence-Only approach, we select the bounding box with the highest confidence score from each model.
Next, we compute the CLIP similarity score between each predicted bounding box and the input text.
The bounding box with the higher CLIP similarity score is then chosen as the final prediction.
\begin{equation}
    CLIPSim_{H} = max(100 * cos(E_{I'_{H}, E_{T}}), 0)
    \label{app:eq:clip_sim_h2f}
\end{equation}
where $E_{I'_{H}}$ is the embedded representation of cropped predicted bounding box of $M_{H}$ with the highest score and $E_{T}$ is the embedded representation of the text input.

\vspace{-4mm}

\begin{equation}
    CLIPSim_{W} = max(100 * cos(E_{I'_{W}, E_{T}}), 0)
    \label{app:eq:clip_sim_e2w}
\end{equation}
where $E_{I'_{W}}$ is the embedded representation of cropped predicted bounding box of $M_{W}$ with the highest score.

\subsection{CLIP-Only (Top-2 + Threshold)}
\label{app:clip_only_top2}
This method is similar to CLIP-Only (Top-1), where we take predictions from both models and compute the CLIP similarity score between the cropped bounding boxes and the input text.
However, unlike the Top-1 variant, we also consider the Top-2 predictions from each model.
However, we do not always include the second predictions; instead, we apply a confidence threshold determined empirically on the validation set, $T=0.95$.
If a model’s second-highest prediction has a confidence score greater than or equal to this threshold, it is included in the comparison.
Finally, we select the bounding box with the highest CLIP similarity score as the final prediction.
The algorithm for this method is presented in \cref{app:alg:clip_top2}.

\begin{algorithm}[t]
\caption{Algorithm of CLIP-Only (Top-1) ensemble method.}
\label{app:alg:clip_only_top1}
\begin{algorithmic}[1]
\Require Top-1 predictions from models $M_{H}$ and $M_{W}$: $(b^{H}, c^{H})$ and $(b^{W}, c^{W})$, input text $t$
\Ensure Final bounding box prediction $b^*$
\State Crop predicted regions from $b^{H}$ and $b^{W}$ to get image patches $I'_{H}$ and $I'_{W}$
\State Compute text embedding $E_T$ from $t$
\State Compute CLIP similarity scores using Eq.~\eqref{app:eq:clip_sim_h2f} and Eq.~\eqref{app:eq:clip_sim_e2w}:
\State \hspace{0.5cm} $s_{H} = \max(100 \cdot \cos(E_{I'_{H}}, E_T), 0)$
\State \hspace{0.5cm} $s_{W} = \max(100 \cdot \cos(E_{I'_{W}}, E_T), 0)$
\If{$s_{H} \geq s_{W}$}
    \State $b^* \leftarrow b^{H}$
\Else
    \State $b^* \leftarrow b^{W}$
\EndIf \\
\Return $b^*$
\end{algorithmic}
\end{algorithm}

\subsection{CLIP Fusion}
\label{app:confidence_clip_fusion}
In this method, we leverage both the CLIP similarity score and the model's confidence score to make the final prediction.
Specifically, we compute a hybrid score for each bounding box by summing its confidence score and its CLIP similarity score.
We use the Top-2 predictions from both models for this process.
However, the CLIP similarity scores and model confidence scores operate on different scales: while confidence scores are normalized in the range $[0,1]$, CLIP similarity scores typically range around $25$.
To balance their influence and prevent the CLIP score from dominating, we scale the CLIP similarity scores by a factor of $0.04$.
The scaled CLIP score is then added to the corresponding confidence score to compute the final hybrid score for each bounding box.
Finally, we select the bounding box with the highest hybrid score as the final prediction.
The algorithm for this method is presented in \cref{app:alg:hybrid_score}.

\begin{algorithm}[t]
\caption{Algorithm for CLIP-Only (Top-2 + Threshold).}
\label{app:alg:clip_top2}
\begin{algorithmic}[1]
\Require  Predictions from models $M_{H}$ and $M_{W}$, confidence threshold $T = 0.95$, input text $t$
\Ensure Final bounding box prediction $b^*$
\State Extract top-2 predictions from each model:
\State \hspace{0.5cm} $P_{H} = \{(b_1^{H}, c_1^{H}), (b_2^{H}, c_2^{H})\}$
\State \hspace{0.5cm} $P_{W} = \{(b_1^{W}, c_1^{W}), (b_2^{W}, c_2^{W})\}$
\State Initialize candidate set $\mathcal{B} = \{b_1^{H}, b_1^{W}\}$
\If{$c_2^{H} \geq T$}
    \State Add $b_2^{H}$ to $\mathcal{B}$
\EndIf
\If{$c_2^{W} \geq T$}
    \State Add $b_2^{W}$ to $\mathcal{B}$
\EndIf
\For{each $b \in \mathcal{B}$}
    \State Compute CLIP similarity score $s_b = \text{CLIP}(b, t)$
\EndFor
\State $b^* = \arg\max_{b \in \mathcal{B}} s_b$ \\
\Return $b^*$
\end{algorithmic}
\end{algorithm}

\begin{algorithm}[t]
\caption{Algorithm of CLIP Fusion method.}
\label{app:alg:hybrid_score}
\begin{algorithmic}[1]
\Require Top-2 predictions from models $M_{H}$ and $M_{W}$, input text $t$, scaling factor $\lambda = 0.04$
\Ensure Final bounding box prediction $b^*$
\State Extract top-2 predictions:
\State \hspace{0.5cm} $P_{H} = \{(b_1^{H}, c_1^{H}), (b_2^{H}, c_2^{H})\}$
\State \hspace{0.5cm} $P_{W} = \{(b_1^{W}, c_1^{W}), (b_2^{W}, c_2^{W})\}$
\State Initialize candidate set $\mathcal{B} = \{(b_1^{H}, c_1^{H}), (b_2^{H}, c_2^{H}), (b_1^{W}, c_1^{W2F}), (b_2^{W}, c_2^{W})\}$
\For{each $(b, c) \in \mathcal{B}$}
    \State Compute CLIP similarity score $s = \text{CLIP}(b, t)$
    \State Compute hybrid score $h = c + \lambda \cdot s$
    \State Store $(b, h)$
\EndFor
\State $b^* = \arg\max_{(b, h)} h$ \\
\Return $b^*$
\end{algorithmic}
\end{algorithm}

\subsection{CLIP-Aware Pointing Ensemble (CAPE)}
\label{app:cape}
In this method, we combine the CLIP-Only (Top-2 + Threshold) and CLIP Fusion approaches.
Through empirical analysis on validation set, we observe that for small objects, the CLIP Fusion method yields more accurate results than other ensemble strategies.
This is primarily because CLIP tends to perform less reliably on smaller objects.
Therefore, we apply CLIP Fusion to not rely on only CLIP-score but also consider the confidence score.
To determine whether an object is small, we follow the definition provided by \cite{chen2021yourefit}: if the object occupies less than $0.48\%$ of the total image area, we classify it as small and use the CLIP Fusion ensemble.
For all other cases, we apply the CLIP-Only (Top-2 + Threshold) method.
The algorithm is presented in \cref{app:alg:CAPE}.

\begin{algorithm}[H]
\caption{CLIP-Aware Pointing Ensemble (CAPE) algorithm.}
\label{app:alg:CAPE}
\begin{algorithmic}[1]
\Require Top-2 predictions from models $M_{H}$ and $M_{W}$, input text $t$, object area threshold $\tau = 0.0048$
\Ensure Final bounding box prediction $b^*$
\State Extract candidate bounding boxes from both models
\State Estimate area ratio $r$ of each candidate bounding box to the image size
\If{$r < \tau$}
    \State $b^* = \text{ConfidenceCLIPFusion}(M_{H}, M_{W}, t)$ \hfill // Refer to Algorithm~\ref{app:alg:hybrid_score}
\Else
    \State $b^* = \text{CLIPOnlyTop2Threshold}(M_{H}, M_{W}, t)$ \hfill // Refer to Algorithm~\ref{app:alg:clip_top2}
\EndIf \\
\Return $b^*$
\end{algorithmic}
\end{algorithm}

\section{Discussion}

\subsection{Why is pointing heatmap useful?}
In ERU, the model must associate a textual instruction with the spatial region indicated by the person in the image.
Without explicit spatial guidance, the model relies on implicit cues, such as hand position or object features, which become ambiguous when multiple candidate objects fit the text instruction.
A pointing line provides a clear directional prior, narrowing the possible targets to regions intersecting or near the line. Formally, if $R(I, T)$ is the set of candidate regions without explicit pointing heatmap ($P$), then pointing guidance:
\begin{equation}
    R'(I, T, P) \subseteq R(I,T)
\end{equation}
where $I$ denotes the image, $T$ refers to the textual description, and $P$ is the pointing line. This reduces ambiguity and helps the model converge to the correct region.

A second benefit is spatial alignment of multimodal embeddings.
Without spatial cues, the latent representation of the intended object may blend with nearby distractors.
The pointing line acts as a spatial attention prior, biasing the model toward regions consistent with the pointing direction.

Finally, using a pointing heatmap also strengthens the gradient signal during training.
The heatmap provides an additional, well-structured supervisory cue, reducing overfitting to spurious text or background correlations.
Effectively, $P$ serves as a spatial inductive bias that stabilizes and accelerates learning.

In summary, the pointing line provides an explicit spatial prior, constrains the candidate space, improves multimodal alignment, and yields more efficient and accurate grounding.

\subsection{Why does CAPE enhance the performance?}
The head-to-finger line provides a global directional prior, while the wrist-to-finger line captures local articulation.
Together, they offer complementary spatial cues, allowing each network to capture different aspects of the pointing geometry and reducing the risk of misalignment across varied pointing poses.
Therefore, we introduce CAPE module to ensemble the outputs of both models ($M_H$ and $M_F$).
We incorporate CLIP in addition to model confidence scores because CLIP learns a joint image–text embedding space where images and text with matching semantics are close.
Formally, for an image region $I^C$ and a text description $T$, CLIP provides a similarity score:
\begin{equation}
    S_{CLIP}(I^C, T) = \omega (\phi_I(I^C), \phi_T(T))
\end{equation}
where $\phi_I$ and $\phi_T$ are image and text encoders, and $\omega$ is cosine similarity.
This score captures semantic alignment.

Relying only on CLIP is insufficient because, while it provides semantic similarity, it does not encode geometric cues such as pointing direction. This can lead to imprecise localization in cluttered scenes or when multiple semantically plausible objects exist. Conversely, using only model confidence is also unreliable: pointing-based networks provide spatial priors but can be overconfident even when predictions are wrong due to occlusion, unusual poses, or network biases.

The two pointing models provide spatial priors, while CLIP provides semantic priors.
Combining them reduces both geometric and linguistic ambiguity.
Conceptually, the fusion functions like a Bayesian ensemble.
The pointing models supply a geometry-driven likelihood over regions, while CLIP acts as a semantic prior.
Integrating these signals produces more sharper and reliable posterior estimate.

As an alternative to CLIP model, we also experimented with SigLIP~\cite{zhai2023sigmoid}, which provides strong semantic grounding. However, replacing CLIP with SigLIP in the CAPE module yielded slightly lower performance (see \cref{app:tab:clip_vs_siglit}), so we continue using CLIP.

In summary, CAPE fuses pointing-based confidence with CLIP similarity, integrating complementary spatial and semantic signals to improve referent prediction in challenging poses.

\begin{table}[]
    \centering
    \footnotesize
    \begin{tabular}{c|ccc}
    \toprule
        Method & IoU=0.25 & IoU=0.50 & IoU=0.75 \\
        \midrule
         CAPE with SigLIP & 74.5 & 65.1 & 35.4 \\
         CAPE with CLIP & 75.0 & 65.4 & 35.7 \\
    \bottomrule
    \end{tabular}
    \caption{We compare SigLIP with CLIP as part of the CAPE module. CLIP demonstrates higher performance than SigLIP. The experiments are conducted on the YouRefIt dataset.}
    \label{app:tab:clip_vs_siglit}
\end{table}

\begin{table}[]
    \centering
    \footnotesize
    %\resizebox{\linewidth}{!}{
    \begin{tabular}{c|c|ccc}
    \toprule
        Method & Heatmap Input & IoU=0.25 & IoU=0.5 & IoU=0.75  \\
        \midrule
         $M_{H}$ & Head-to-finger  & 72.9 & 62.3 & 35.1 \\ 
         $M_{H}$ & Wrist-to-finger & 73.0 & 62.0 & 35.6 \\
         $M_{H}$ & Ground-Truth & 73.0 & 62.1 & 36.2 \\
         $M_{H}$ & None & 71.6 & 60.7 & 32.9 \\
         \midrule
         $M_{W}$ & Wrist-to-finger & 69.6 & 60.7 & 31.5 \\
         $M_{W}$ & Head-to-finger  & 68.1 & 59.2 & 28.4 \\
         $M_{W}$ & Ground-Truth & 69.7 & 59.9 & 31.9 \\
         $M_{W}$ & None & 68.8 & 59.7 & 31.2 \\
         \bottomrule
    \end{tabular}%}
    \caption{We analyze the performance of our models under different heatmap inputs during inference.}
    \label{app:tab:robustness_test_heatmap}
\end{table}

\subsection{Effect of Heatmap}

We evaluate the robustness of our models ($M_{H}$ and $M_{W}$) using different heatmap inputs (Table \ref{app:tab:robustness_test_heatmap}). First, we test $M_H$ using a wrist-to-finger heatmap. While IoU 0.25 and 0.75 show slight gains, integrating this into CAPE reduces overall performance, indicating a loss of complementary effectiveness.
In our main experiments, we generate both heatmaps using predicted keypoints from a pose estimation model.
In contrast, the \textit{Ground-Truth} setup uses the annotated pointing coordinates, including the center of the ground-truth bounding box, to generate the heatmap.
As expected, this yields a small performance boost since the line is guaranteed to intersect the target object’s center. However, the improvement is marginal, indicating that our predicted heatmaps are already accurate and that the pointing line reliably guides the model.
Finally, providing a zero heatmap (\textit{None}) removes explicit pointing guidance. Performance drops but remains reasonable, demonstrating that the model can rely on text and visual context. Notably, since the model is trained with explicit pointing heatmaps, it has learned to interpret pointing cues; therefore, even without a heatmap at inference, it can still achieve meaningful performance.
Repeating the same experiments for $M_W$ shows similar trends, reinforcing the complementary nature of the two heatmaps.

\begin{table}[]
    \centering
    %\resizebox{\linewidth}{!}{
    \footnotesize
    \begin{tabular}{c|c|ccc}
    \toprule
        Method & Text & IoU=0.25 & IoU=0.5 & IoU=0.75  \\
        \midrule
         $M_{H}$ & Original  & 72.9 & 62.3 & 35.1 \\ 
         $M_{H}$ & Dummy     & 34.0 & 26.5 & 12.3  \\
         $M_{H}$ & Random    & 24.9 & 19.0  & 9.4 \\
         $M_{H}$ & No Text    & 39.2 & 27.9  & 10.2 \\
         \midrule
         $M_{W}$ & Original & 69.6 & 60.7 & 31.5 \\
         $M_{W}$ & Dummy  & 34.6 & 25.5 & 11.0 \\
         $M_{W}$ & Random  & 25.4 & 19.3 & 9.24 \\
         $M_{W}$ & No Text  & 43.3 & 31.1 & 11.7 \\
         \bottomrule
    \end{tabular}%}
    \caption{We analyze the performance of our models under different text inputs during inference.}
    \label{app:tab:robustness_test_text}
\end{table}

\subsection{Effect of Text Input}

In summary, for the embodied reference understanding task, text serves as the primary cue for object identification, while pointing information, derived from both the input image and the heatmap, plays a complementary, assistive role.

In Table \ref{app:tab:robustness_test_text}, we evaluate our models under varying text input conditions to assess the role of textual guidance. 
The \textit{Original} setup uses the dataset annotations. 
In the \textit{Dummy} scenario, the text is replaced with the generic word “object,” while in the \textit{Random} scenario, text is randomly sampled from other test examples, sometimes contradicting the actual target. Both scenarios cause a notable performance drop, highlighting the importance of textual input. Without accurate text instructions, the pointing line often intersects multiple objects, making it insufficient to identify the correct referent in cluttered scenes.

In summary, text provides the primary cue for referent identification in ERU, while pointing information from the image and heatmap serves as a complementary signal.

\subsection{Finetuning or Frozen Text Encoder?}
\label{app:frozen_text_encoder}
\begin{figure}
    \centering
    \includegraphics[width=1\linewidth]{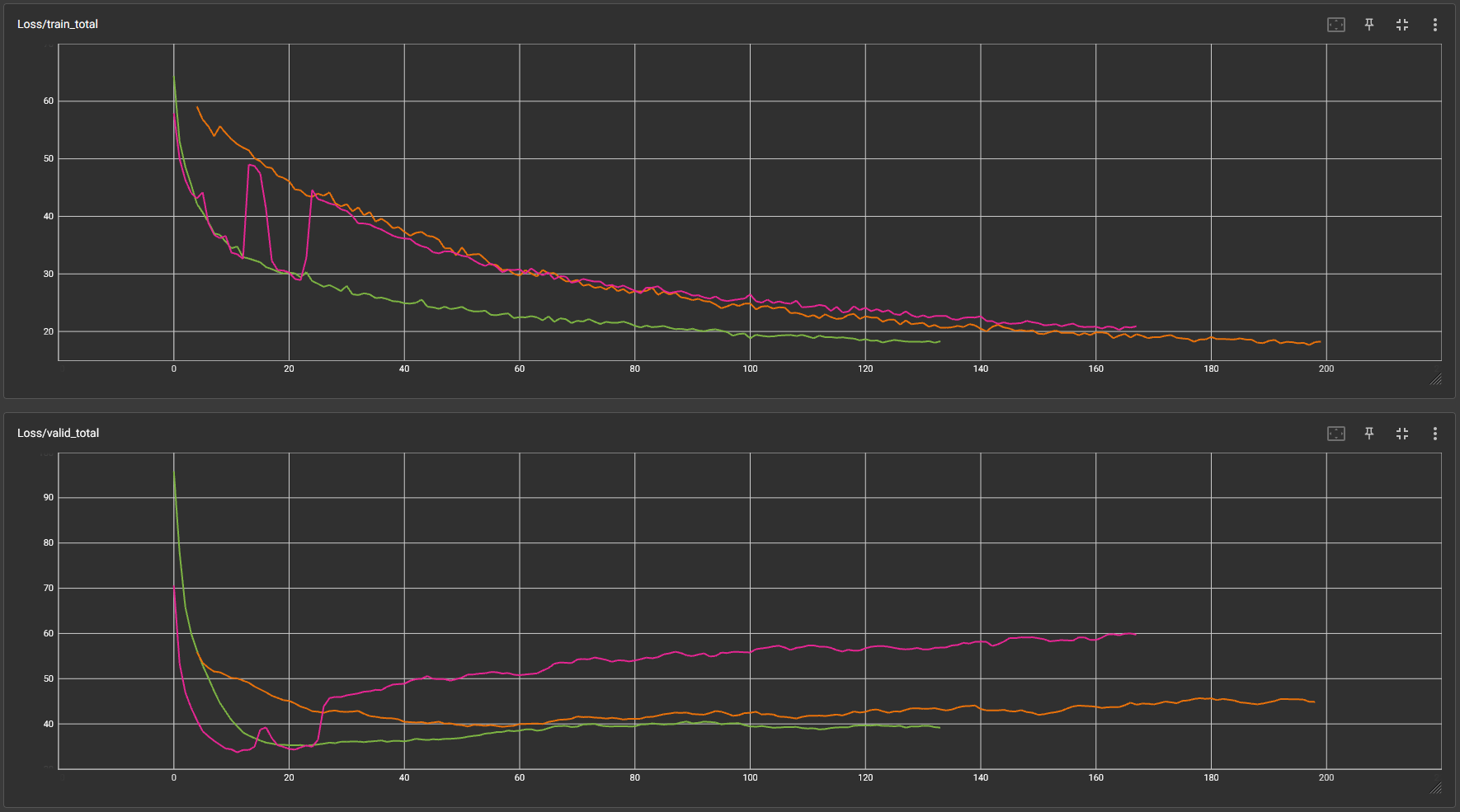}
    \caption{Graphs of training and validation loss from three separate runs are shown. The green and pink lines represent training with a finetuned text encoder, while the orange line corresponds to training with a frozen text encoder.}
    \label{app:fig:loss1_for_frozen_text_encoder}
\end{figure}

We investigate the impact of freezing the text encoder during training. 
Based on multiple training experiments, we observed that finetuning the text encoder sometimes leads to instability, represented by the pink line in \cref{app:fig:loss1_for_frozen_text_encoder}. 
In contrast, freezing the text encoder results in consistently stable training, as shown by the green and orange lines in \cref{app:fig:loss1_for_frozen_text_encoder}. 
%Therefore, achieving stable training with a finetuned text encoder requires several attempts.
This indicates that finetuning introduces optimization sensitivity and may require multiple attempts to obtain a stable run.
%However, despite this challenge, finetuning the text encoder ultimately yields better performance.
Despite this instability, finetuning ultimately yields better results.
%This suggests that adapting the text features to the specific task of embodied reference understanding provides a meaningful benefit.
Table~\ref{tab:ablation_text_encoder} shows that, for both Setup \texttt{A} and Setup \texttt{F}, finetuning improves mAP at IoU 0.25 and 0.5, and also provides better $C_D$ scores.
These gains suggest that adapting the text representations to the specific semantics of embodied reference understanding provides a meaningful advantage, even though the training process becomes less stable.

\begin{table}[t]
    \centering
    %\tiny
    \resizebox{\linewidth}{!}{\begin{tabular}{lccccc}
        \toprule
        Text Encoder & IoU=0.25 & IoU=0.5 & IoU=0.75 & CLIP $\uparrow$ & $C_D$ $\downarrow$ \\
        \midrule
        Frozen, Setup \texttt{A} & 69.8 & 59.5 & \textbf{33.1} & 0.2454 & 0.2994 \\
        Finetuned, Setup \texttt{A} & \textbf{71.2} & \textbf{60.1} & 32.8 & \textbf{0.2469} & \textbf{0.2662} \\
        \midrule
        Frozen, Setup \texttt{F}  & 71.3 & 62.1 & 33.7 & \textbf{0.2460} & 0.2634 \\
        Finetuned, Setup \texttt{F} & \textbf{72.9} & \textbf{62.2} & \textbf{35.1} & 0.2457 & \textbf{0.2490} \\
        \bottomrule
    \end{tabular}}
    \caption{Ablation study for frozen and finetuned text encoder. }
    \label{tab:ablation_text_encoder}
\end{table}

%\vspace{-4mm}

\begin{table}[h!]
    \centering
    \begin{tabular}{c|c}
    \toprule
    Name & Value \\
    \midrule
    Learning rate vision backbone &  $5e^-5$ \\
    Learning rate - text backbone &  $1e^-4$ \\
    Learning rate & $5e^-5$ \\
    $\lambda_1$ & 2 \\
    $\lambda_2$ & 1 \\
    $\lambda_3$ & 10 \\
    $\lambda_4$ & 10 \\
    $\lambda_5$ & 1 \\
    $\lambda_6$ & 1 \\
    Transformer encoder layer & 6 \\
    Transformer decoder layer & 6 \\
    Number of attention heads per layer & 8 \\
    MLP dimension & 2048 \\
    Dropout & p = 0.1 \\
    Weight decay & 1e-4 \\
    Batch size & 4 \\
    Text encoder & Roberta-Base \\
    Image encoder & ResNet-101 \\
    Heatmap encoder & ResNet-18 \\
    Position embedding & Sine \\
    Input dim of transformer encoder & 256 \\
    Number of queries & 20 \\
    \bottomrule
    \end{tabular}
    \caption{Hyperparameters and other training details.}
    \label{app:hyperparameters_table}
\end{table}

\section{Hyperparameters and Other Training Parameters}
\label{app:hyperparameters}
In \cref{app:hyperparameters_table}, we present empirically determined hyperparameters and other key training details to support reproducibility.
For most hyperparameters, we follow established values from the literature, as they have been thoroughly analyzed in prior work~\cite{li2023understanding,carion2020end,li2022dn,meng2021conditional,zhang2022dino,zhu2020deformable,liu2024grounding}.
We also observe that training is not sensitive to the choice of loss coefficients, as the resulting accuracy differences are negligible.

\section{More Results}
\label{app:more_results}

\begin{table*}
    \centering
    %\resizebox{\textwidth}{!}{
    \footnotesize
    \begin{tabular}{l|cccc|cccc|cccc}
        \toprule
        IoU Threshold for mAP & \multicolumn{4}{c|}{0.25} & \multicolumn{4}{c|}{0.50} & \multicolumn{4}{c}{0.75} \\
        \midrule
        Object Sizes & \multicolumn{1}{c}{All} & \multicolumn{1}{c}{S} & \multicolumn{1}{c}{M} & \multicolumn{1}{c|}{L} & \multicolumn{1}{c}{All} & \multicolumn{1}{c}{S} & \multicolumn{1}{c}{M} & \multicolumn{1}{c|}{L} & \multicolumn{1}{c}{All} & \multicolumn{1}{c}{S} & \multicolumn{1}{c}{M} & \multicolumn{1}{c}{L} \\
        \midrule
        PaliGemma2     & 54.4 & 46.2 & 53.0 & 64.1 & \textbf{51.2} & 42.4 & 47.2 & 64.1 & \textbf{44.5} & 26.2 & 43.3 & \textbf{64.1} \\
        Qwen2.5vl      & 36.4 & 21.0 & 37.2 & 51.1 & 31.6 & 14.9 & 32.0 & 47.9 & 21.0 & 10.0 & 19.3 & 33.7 \\
        Grounding DINO & 47.6 & 40.7 & 47.9 & 54.4 & 46.8 & 38.7 & 47.4 & 54.4 & \textbf{44.5} & \textbf{33.1} & \textbf{46.1} & 54.4 \\
        \midrule
        Touch-Line-EWL & 50.7 & 49.8 & 53.2 & 45.6 & 46.5 & 45.5 & 49.3 & 41.3 & 24.6 & 20.3 & 30.9 & 29.3 \\
        Touch-Line-VTL & 45.9 & 44.2 & 51.2 & 32.6 & 40.3 & 39.5 & 43.2 & 32.6 & 20.6 & 13.9 & 29.4 & 32.6 \\
        \midrule
        $M_{H}$ & \textbf{55.5} & 52.3	& \textbf{57.6}	& 70.6	& 46.4	& 39.7	& \textbf{55.6}	& 70.7 & 15.2 & 3.3 & 26.8 & 62.0 \\
        $M_{W}$ & 48.6 & 46.2	& 49.4	& 64.1	& 42.5 & 40.1 & 43.1 & 59.8 & 17.7 & 7.8 & 28.8 & 43.5 \\
        CAPE w/ $M_H$ + $M_W$ & \textbf{55.5} & \textbf{52.9} & 55.8 & \textbf{75.0} & 50.5 & \textbf{47.3} & 50.9 & \textbf{75.0} & 18.6 & 9.1 & 26.6 & 57.6 \\
        \bottomrule
    \end{tabular}%}
    \caption{Comparison of our model with prior works, SOTA LMMs, and Grounding-DINO in terms of mean Average Precision (mAP) at different IoU thresholds, across various object sizes, on the CAESAR dataset.}
    \label{app:tab:comparison_iou_CAESAR}
\end{table*}

In \cref{app:tab:comparison_iou_CAESAR}, we report the results of our model on the CAESAR dataset, which is simulation-based. 
We compare our model with SOTA methods, including Touch-Line, GroundingDINO, PaliGemma2, and Qwen2.5vl. 
Despite the significant domain gap, our model achieves very accurate performance, attaining SOTA scores in most cases. 
Under the IoU threshold of 0.75, PaliGemma2 and GroundingDINO achieve higher scores.

In \cref{app:tab:comparison_iou}, we report the mAP at three different thresholds for all ablation setups, categorized by object size.
Similarly, in \cref{app:tab:comparison_clip}, we present the CLIP scores and $C_D$ metrics for the same setups, also with respect to object size.

\begin{figure*}
    \centering
    \includegraphics[width=1\linewidth]{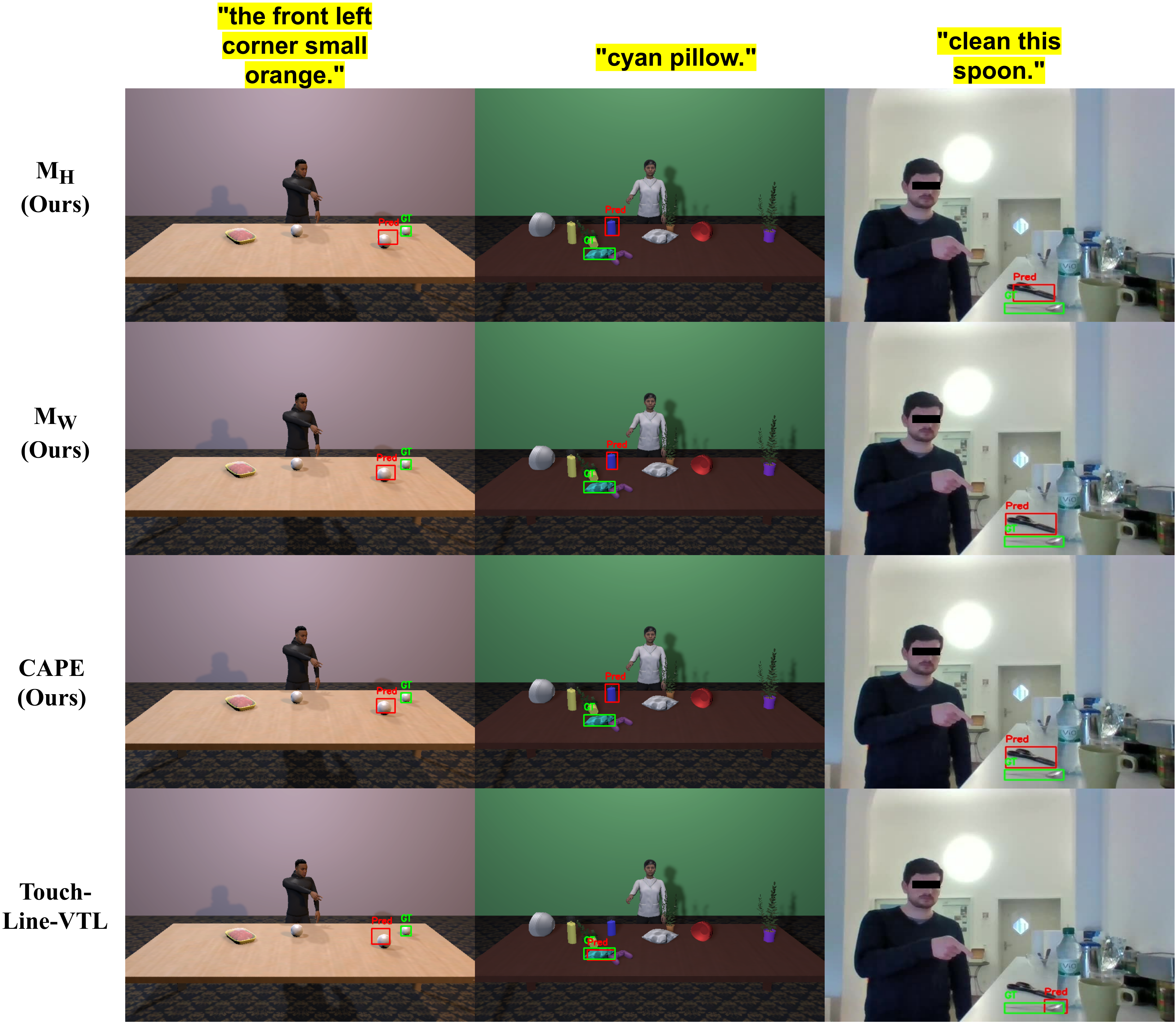}
    \caption{Some examples for failure cases from ISL pointing and CAESAR datasets. }
    \label{app:fig:failures}
\end{figure*}

In \cref{app:fig:failures}, we present example failure cases from the ISL Pointing and CAESAR datasets. In the first column, all models predict the same object, as it aligns better with the pointing line and the text provides limited semantic guidance. Localization-related text is particularly challenging for the models to interpret. In the second column, only Touch-Line-VTL predicts the correct object. In the third column (from ISL Pointing), our models detect the scissors correctly, whereas Touch-Line-VTL detects only a small part of the spoon. Due to perspective and the lack of depth information, our models align the scissors with the pointing line.

\begin{figure*}
    \centering
    \includegraphics[width=1\linewidth]{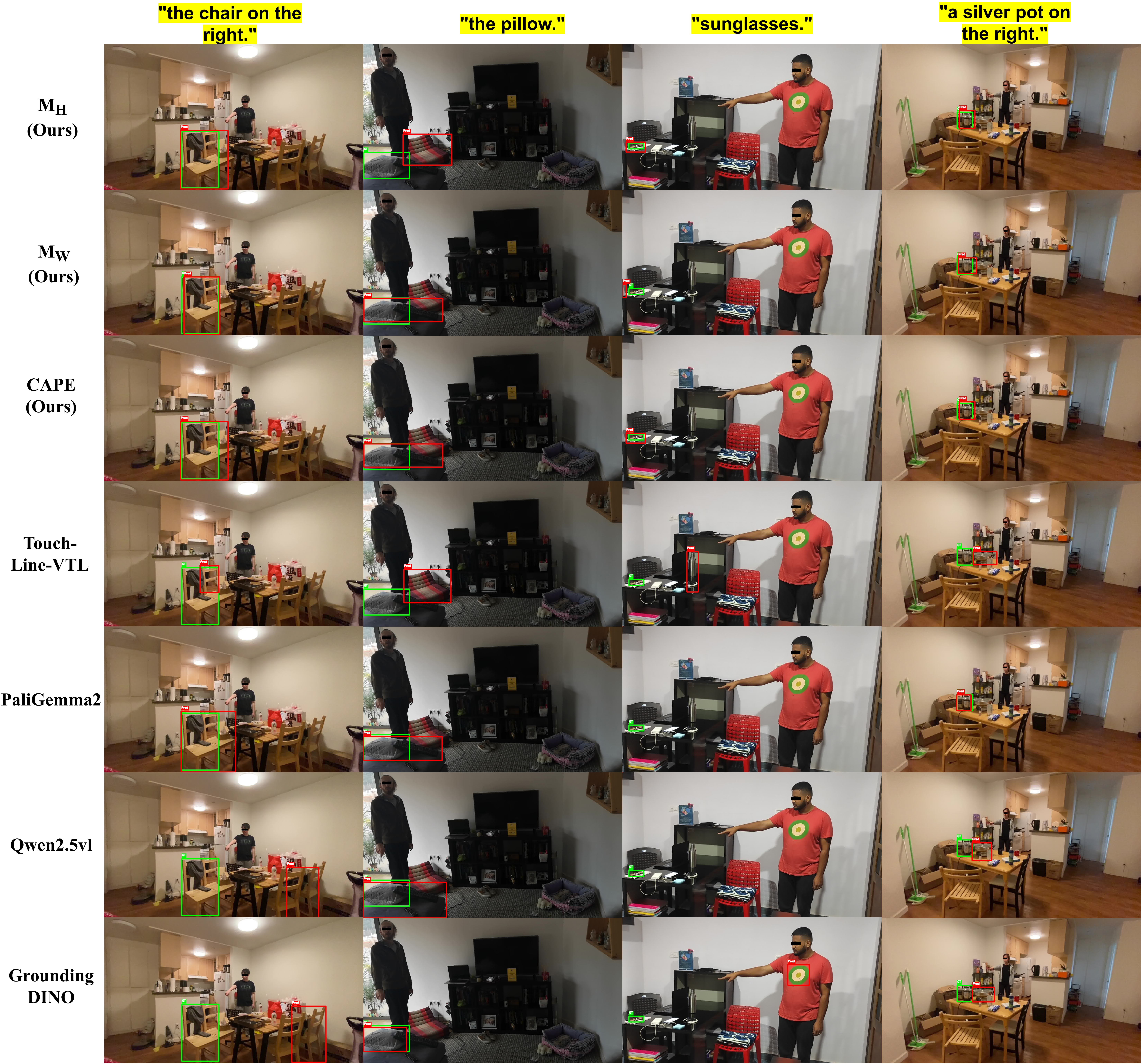}
    \caption{Example results from different models on the YouRefIt dataset.}
    \label{app:fig:example_yourefit_supp}
\end{figure*}

In \cref{app:fig:example_yourefit_supp}, we present additional example images from the YouRefIt dataset and qualitatively compare our model with SOTA models. 
CAPE achieves high accuracy in most cases, whereas Touch-Line often fails to detect the correct object, and the LMMs largely ignore the pointing cue, relying only on the text instruction.

\begin{figure*}
    \centering
    \includegraphics[width=1\linewidth]{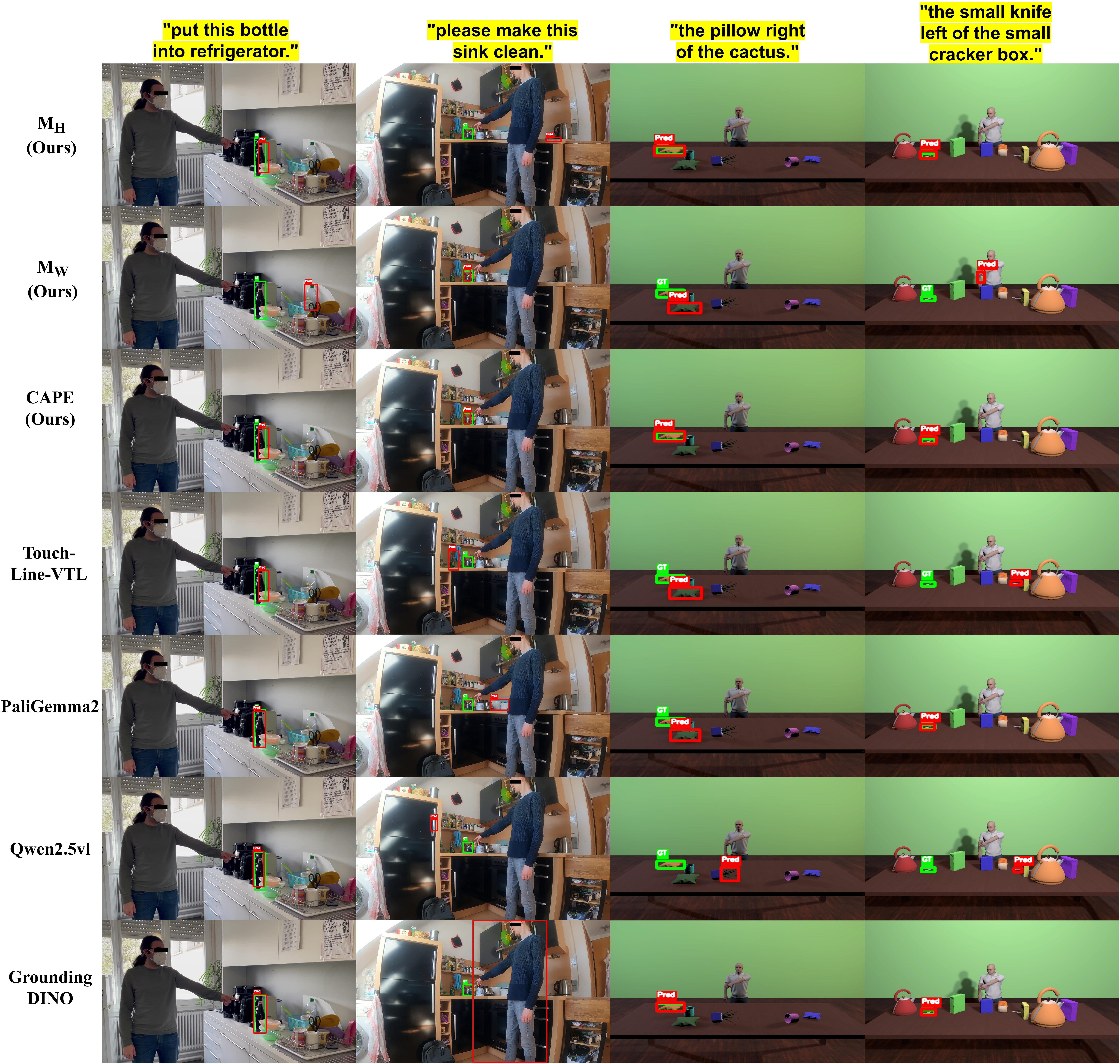}
    \caption{Example results from different models on the ISL Pointing and CAESAR datasets. While first two columns have examples from ISL Pointing dataset~\cite{constantin2022interactive}, remaining two columns contain sample images from CAESAR dataset~\cite{islam2022caesar}.}
    \label{app:fig:example_ISL_and_CAESAR}
\end{figure*}

In \cref{app:fig:example_ISL_and_CAESAR}, we present example images from the ISL Pointing and CAESAR datasets. 
The first two columns show ISL Pointing images, while the remaining two columns contain CAESAR samples, which are simulation-based. 
In the first ISL example, all models except $M_W$ predict the object correctly. 
$M_W$ makes wrong prediction because its prediction aligns better with the wrist-to-fingertip pointing line than the GT object, and the provided text does not conflict with this prediction. 
In the second example, $M_H$ fails, but CAPE correctly selects $M_W$’s prediction. 
Touch-Line-VTL, PaliGemma2, Qwen2.5vl, and GroundingDINO make mistakes as they overlook the pointing cue.
On the CAESAR dataset, despite the severe domain gap, our model and Touch-Line-VTL achieve strong performance. 
Our models closely follow the pointing cue and detect the correct object more accurately than other models. 
Given the challenging text instructions in both datasets, accurately and efficiently leveraging the pointing direction is key, explaining why CAPE surpasses all other models.

\begin{table*}
    \centering
    \resizebox{\textwidth}{!}{
    \begin{tabular}{c|l|c|c|cccc|cccc|cccc}
        \toprule
        & IoU Threshold for mAP & & & \multicolumn{4}{c}{0.25} & \multicolumn{4}{|c}{0.50} & \multicolumn{4}{|c}{0.75} \\
        %\cmidrule(lr){2-5} \cmidrule(lr){6-9} \cmidrule(lr){10-13}
        \midrule
        Setup & Object Sizes & Heatmap Injection & Ensemble & \multicolumn{1}{c}{All} & \multicolumn{1}{c}{S} & \multicolumn{1}{c}{M} & \multicolumn{1}{c}{L} & \multicolumn{1}{|c}{All} & \multicolumn{1}{c}{S} & \multicolumn{1}{c}{M} & \multicolumn{1}{c}{L} & \multicolumn{1}{|c}{All} & \multicolumn{1}{c}{S} & \multicolumn{1}{c}{M} & \multicolumn{1}{c}{L} \\
        \midrule
        & PaliGemma2     & - & - & 58.8 & 29.0 & 53.5 & 75.8 & 46.9 & 22.1 & 50.8 & 68.0 & 31.7 & 6.2 & 34.1 & 54.8 \\
        & Qwen2.5vl-3b   & - & - & 38.9 & 17.0 & 41.8 & 58.0 & 31.0 & 11.1 & 33.6 & 48.1 & 20.0 & 5.7 & 19.8 & 34.5 \\
        & Grounding DINO & - & - & 57.9 & 38.0 & 60.9 & 74.9 & 54.9 & 35.7 & 59.3 & 69.6 & \textbf{42.3} & \textbf{22.7} & \textbf{45.9} & \textbf{58.4} \\
        \midrule
        & Touch-Line-EWL \cite{li2023understanding} & - & - & {69.5} & {56.6} & {71.7} & 80.0 & 60.7 & {44.4} & 66.2 & 71.2 & 35.5 & {11.8} & {38.9} & 55.0 \\
        & Touch-Line-VTL \cite{li2023understanding} & - & - & 71.1 & 55.9 & 75.5 & {81.7} & {63.5} & 47.0 & {70.2} & {73.1} & \underline{39.0} & 13.4 & \underline{45.2} & \underline{57.8} \\
        \midrule
        \texttt{1} & Baseline & - & - & 71.2 & 59.8 & 73.0 & 80.9 &	60.1 &	43.2 &	66.4 &	70.5 &	32.8 &	8.6 &	35.4 &	53.4 \\
        \texttt{2} & Setup \texttt{1} + object center prediction & - & - & 70.8 & 59.5 & 73.2 & 79.5 & 61.7	& 45.8 &	67.4	& 71.5 &	34.6 &	\underline{14.0} &	38.1 &	51.1 \\
        \texttt{3} & Setup \texttt{1} + W2F heatmap & Embedding feature & - & 68.9 & 55.5 &	74.0 &	77.4 &	59.7 &	42.0 &	68.1 &	69.3 &	32.4 &	8.7 &	37.8 &	50.6 \\
        \texttt{4} & Setup \texttt{1} + H2F heatmap & Embedding feature & - & 71.9 & 57.5 &	74.1 &	84.1 &	62.8 &	43.3 &	70.1 &	75.1 &	33.8 & 10.0 &	35.6 & 55.3 \\
        \texttt{5} & Setup \texttt{2} + W2F heatmap & Embedding feature & - & 69.6	& 57.7 & 74.9 & 76.4 & 60.7 & 45.5 &	69.0 &	68.2 &	31.5 &	12.1 &	35.3 &	47.0 \\
        \texttt{6} & Setup \texttt{2} + H2F heatmap & Embedding feature & - & 72.9 & 59.5 &	77.8 &	81.4 &	62.3 &	44.3 &	70.9 &	42.0 &	35.1 &	10.4 &	39.3 &	55.3 \\
        \texttt{7} & Setup \texttt{1} + W2F heatmap + H2F heatmap & Embedding feature & - & 70.2 &	59.0 &	71.4 &	80.2 &	60.2 &	44.5 &	65.6 &	70.5 &	33.8 & 11.9 &	39.3 &	50.0 \\
        \texttt{8} & Setup \texttt{6} & Channel-wise input & - & 68.7 &	58.5 &	71.9 &	75.6 &	58.5 &	44.8 &	65.0 &	65.9 &	33.4 &	13.9 &	37.4 &	48.6 \\
        \texttt{9} & Setup \texttt{6} & Channel-wise feature & - & 72.6	& 60.0 &	77.5 &	80.4 &	58.4 &	38.8 &	68.7 &	68.1 &	27.9 &	8.5 &	32.6 &	42.5 \\
        \texttt{10} & Setup \texttt{5} + Setup \texttt{6} & Embedding feature & Confidence-Only & 73.4 &	62.7 & 78.6 &	79.0 &	64.1 &	49.0 &	72.7 &	71.0 &	34.1 &	13.4 &	38.2 &	50.5 \\ 
        \texttt{11} & Setup \texttt{5} + Setup \texttt{6} & Embedding feature & CLIP-Only (Top-1) & 73.5 & 59.7 &	79.9 &	81.2 &	64.2 &	46.5 &	74.3 &	72.2 &	35.4 &	12.7 &	39.8 &	53.4 \\
        \texttt{12} & Setup \texttt{5} + Setup \texttt{6} & Embedding feature & CLIP-Only (Top-2 + Threshold) & 73.8 &	59.7 &	80.2 &	81.6 &	64.1 &	46.3 &	74.3 &	72.2 &	35.5 &	12.4 &	40.1 &	53.9  \\
        \texttt{13} & Setup \texttt{5} + Setup \texttt{6} & Embedding feature & CLIP Fusion & 73.6 &	63.2 &	78.4 &	79.6 &	64.4 &	49.3 &	72.8 &	71.7 &	34.5 &	13.4 &	38.5 &	51.7 \\
        \texttt{14} & Setup \texttt{5} + Setup \texttt{6} & Embedding feature & CAPE & 75.0 &	63.2 &	80.2 &	81.8 &	65.4 &	49.5 &	74.3 &	72.7 & 35.7 &	13.4 &	40.1 &	53.4 \\
        \texttt{15} & Setup \texttt{6} w/ frozen text encoder & Embedding feature & - &  71.3 &	56.0 &	74.9 &	83.1 &	62.1 &	42.5 &	69.5 &	74.4 &	33.7 &	10.2 &	35.8 &	54.6 \\
        \texttt{16} & Setup \texttt{6} w/ Conic Attention Heatmap ($15^{\circ}$) & Embedding feature & - & 70.0 & 54.2 & 72.2 &	83.3 &	60.0 &	42.0 &	66.6 &	71.5 &	33.9 &	12.2 &	40.6 &	48.8 \\
        \texttt{17} & Setup \texttt{6} w/ Conic Attention Heatmap ($30^{\circ}$) & Embedding feature & - & 71.5 &	58.7 &	73.5 &	82.1 &	59.4 &	41.8 &	64.7 &	71.7 &	30.8 &	7.0 &	32.4 &	52.7 \\
        \midrule
        & Ours - Final & Embedding feature & CAPE & \textbf{75.0} & \textbf{63.2} & \textbf{80.2} & \textbf{81.8} & \textbf{65.4} & \textbf{49.5} & \textbf{74.3} & \textbf{72.7} & {35.7} & {13.4} & {40.1} & {53.4} \\
        \bottomrule
    \end{tabular}}
    \caption{Detailed mAP results of all ablation setups with respect to the object size.}
    \label{app:tab:comparison_iou}
\end{table*}

\begin{table*}[t]
    \centering
    \resizebox{\textwidth}{!}{
    %\footnotesize
    \begin{tabular}{clcccccccccc}
        \toprule
        & & & & \multicolumn{4}{c}{CLIP Score} & \multicolumn{4}{c}{$C_D$}  \\
        %\cmidrule(lr){2-5} \cmidrule(lr){6-9} \cmidrule(lr){10-13}
        \midrule
        Setup & Object Sizes & Heatmap Injection & Ensemble & \multicolumn{1}{c}{All} & \multicolumn{1}{c}{S} & \multicolumn{1}{c}{M} & \multicolumn{1}{c}{L} & \multicolumn{1}{c}{All} & \multicolumn{1}{c}{S} & \multicolumn{1}{c}{M} & \multicolumn{1}{c}{L}  \\
        \midrule
        & Touch-Line-EWL \cite{li2023understanding} & - & - & 0.2456 & 0.2312 & 0.2435 & 0.2615 &	0.3168 &	0.3006 &	0.2903 &	0.3564 \\
        & Touch-Line-VTL \cite{li2023understanding} & - & - & 0.2456 & 0.2308 & 0.2440 & 0.2615 &	0.2843 &	0.2809 &	0.2276 &	0.3393 \\
        \midrule
        \texttt{1} & Baseline & - & - & 0.2470 & 0.2334 & 0.2447 & 0.2620 & 0.2662	& 0.2182 &	0.2506 &	0.3266 \\
        \texttt{2} & Setup \texttt{1} + object center prediction & - & - & 0.2446 & 0.2316 & 0.2426 & 0.2596 &	0.2707 &	0.2201 &	0.2692 &	0.3222 \\
        \texttt{3} & Setup \texttt{1} + W2F heatmap & Embedding feature & - & 0.2451 & 0.2311 & 0.2432 & 0.2604 &	0.2984 &	0.2763 &	0.2616 &	0.3533 \\
        \texttt{4} & Setup \texttt{1} + H2F heatmap & Embedding feature & - & 0.2458 & 0.2312 & 0.2444 & 0.2617 &	0.2694 &	0.2649 &	0.2233 &	0.3172  \\
        \texttt{5} & Setup \texttt{2} + W2F heatmap & Embedding feature & - & 0.2448 & 0.2313 & 0.2435 & 0.2593 & 	0.2770 &	0.2568 &	0.2319 &	0.3382 \\ 
        \texttt{6} & Setup \texttt{2} + H2F heatmap & Embedding feature & - & 0.2458 & 0.2318 & 0.2448 & 0.2606 &	0.249 &	0.222 &	0.2202 &	0.3024 \\
        \texttt{7} & Setup \texttt{1} + W2F heatmap + H2F heatmap & Embedding feature & - & 0.2448 & 0.2322 & 0.2426 & 0.2593 &	0.2744 &	0.2540 &	0.2314 &	0.3335 \\ %*
        \texttt{8} & Setup \texttt{6} & Channel-wise input & - & 0.2449 & 0.2311 & 0.2424 & 0.2609 & 0.2959 & 0.2437 & 0.2642 & 0.3767  \\
        \texttt{9} & Setup \texttt{6} & Channel-wise feature & - & 0.2457 & 0.2318 & 0.2437 & 0.2613 & 0.2598 & 0.2289 & 0.2119 & 0.3339 \\
        \texttt{10} & Setup \texttt{5} + Setup \texttt{6} & Embedding feature & Confidence-Only &  0.2578 & 0.2480 & 0.2614 & 0.2641 &	0.247 &	0.2176 &	0.2042 &	0.3141  \\ 
        \texttt{11} & Setup \texttt{5} + Setup \texttt{6} & Embedding feature & CLIP-Only (Top-1) & 0.2644 & 0.2563 & 0.2670 & 	\textbf{0.2699} &	\textbf{0.2391} &	0.2242 &	0.2011 &	\textbf{0.2878}  \\
        \texttt{12} & Setup \texttt{5} + Setup \texttt{6} & Embedding feature & CLIP-Only (Top-2 + Threshold) & 0.2460 & \textbf{0.2656} & 0.2448 & 0.2606 &	0.2422 &	0.2158 &	\textbf{0.2009} &	0.3060  \\
        \texttt{13} & Setup \texttt{5} + Setup \texttt{6} & Embedding feature & CLIP Fusion & 0.2581 & 0.2642 & 0.2485 & 0.2614 &	0.2434 &	0.2137 &	0.2028 &	0.3090 \\
        \texttt{14} & Setup \texttt{5} + Setup \texttt{6} (Final Model) & Embedding feature & CAPE & \textbf{0.2661} & 0.2642 & \textbf{0.2672}	& 0.2670 &	0.2476 &	\textbf{0.2137} &	0.2241 &	0.3023 \\
        \texttt{15} & Setup \texttt{6} w/ frozen text encoder & Embedding feature & - & 0.2461 & 0.2318 & 0.2441 & 0.2618 &	0.2634 &	0.2626 &	0.2233 &	0.3006  \\
        \texttt{16} & Setup \texttt{6} w/ Conic Attention Heatmap ($15^{\circ}$) & Embedding feature & - & 0.2462 & 0.2321 & 0.2442 & 0.2620 & 0.2695 & 0.2685 & 0.2344 & 0.3028 \\
        \texttt{17} & Setup \texttt{6} w/ Conic Attention Heatmap ($30^{\circ}$) & Embedding feature & - & 0.2464 & 0.2323 & 0.2444 & 0.2621 &	0.2765 &	0.2750 &	0.2499 &	0.3026 \\
        \midrule
        & Ours - Final & Embedding feature & CAPE & \textbf{0.2661} &	0.2642 &	\textbf{0.2672} &	0.2670 &	0.2476 &	\textbf{0.2137} &	0.2241 &	0.3023 \\
        \bottomrule
    \end{tabular}}
    \caption{Detailed CLIP scores and $C_D$ metrics of all ablation setups with respect to the object size.}
    \label{app:tab:comparison_clip}
\end{table*}

